\documentclass[lettersize,journal]{IEEEtran}
\usepackage{amsmath,amsfonts}
\usepackage{algorithmic}
\usepackage{algorithm}
\usepackage{array}
\usepackage{textcomp}
\usepackage{stfloats}
\usepackage{url}
\usepackage{verbatim}
\usepackage{graphicx}

\usepackage{subcaption}
\usepackage{wrapfig}
\usepackage{multirow}
\usepackage{amsmath}
\usepackage{amssymb}

\usepackage[utf8]{inputenc} % allow utf-8 input
\usepackage[T1]{fontenc}    % use 8-bit T1 fonts
\usepackage{hyperref}       % hyperlinks
\usepackage{url}            % simple URL typesetting
\usepackage{booktabs}       % professional-quality tables
\usepackage{amsfonts}       % blackboard math symbols
\usepackage{nicefrac}       % compact symbols for 1/2, etc.
\usepackage{microtype}      % microtypography
\usepackage{xcolor}    

\usepackage[export]{adjustbox}
\usepackage{array}
\usepackage{tabularx}

\usepackage{mwe}
\usepackage{graphbox}

\usepackage{pifont}
\newcommand{\cmark}{\ding{51}}%
\newcommand{\xmark}{\ding{55}}%

\usepackage{balance}

\usepackage[frozencache,cachedir=minted]{minted}

\def\BibTeX{{\rm B\kern-.05em{\sc i\kern-.025em b}\kern-.08em
    T\kern-.1667em\lower.7ex\hbox{E}\kern-.125emX}}

\begin{document}

\title{Dimension Mixer: Group Mixing of Input Dimensions for Efficient Function Approximation}
\author{Suman Sapkota,
Binod Bhattarai

% \thanks{This work has been submitted to the IEEE for possible publication. Copyright may be transferred without notice, after which this version may no longer be accessible.}

\thanks{Suman Sapkota did this work while at NAAMII, Nepal}
\thanks{Binod Bhattarai is with the School of Natural and Computing Sciences at the University of Aberdeen, Aberdeen, UK}
}

% \markboth{Preprint}%
% {Dimension Mixer: Group Mixing of Input Dimensions for Efficient Function Approximation}

\maketitle
\begin{abstract}

The recent success of multiple neural architectures like CNNs, Transformers, and MLP-Mixers motivated us to look for similarities and differences between them. 
We found that these architectures can be interpreted through the lens of a general concept of dimension mixing.
Research on coupling flows and the butterfly transform shows that partial and hierarchical signal mixing schemes are sufficient for efficient and expressive function approximation. 
In this work, we study group-wise sparse, non-linear, multi-layered and learnable mixing schemes of inputs and find that they are complementary to many standard neural architectures.
Following our observations and drawing inspiration from the Fast Fourier Transform, we generalize Butterfly Structure to use non-linear mixer function allowing for MLP as mixing function called Butterfly MLP. We were also able to sparsely mix along sequence dimension for Transformer-based architectures called Butterfly Attention. 
Experiments on CIFAR and LRA datasets demonstrate that the proposed Non-Linear Butterfly Mixers are efficient and scale well when the host architectures are used as mixing function.
Additionally, we propose Patch-Only MLP-Mixer for processing spatial 2D signals demonstrating a different dimension mixing strategy.

\end{abstract}


\section{Introduction}

\begin{figure}
    \centering
    \begin{center}
    \includegraphics[trim=0cm 0cm 0cm 0cm, clip, width=0.99\linewidth]{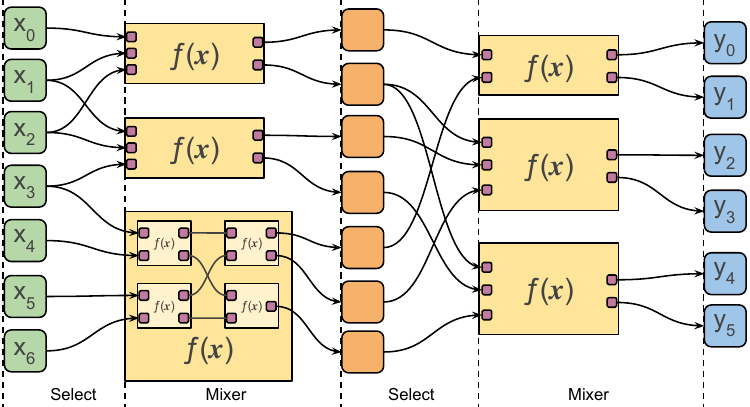}
    \caption{
    An example of a general Dimension Mixer model, split into multiple layers of (i) \emph{Select} and (ii) \emph{Mix} stages. 
    The Select stage selects input dimensions for each of the Mixer units.
    The Mix stage processes inputs and is learned via optimization. 
    This example shows that the mixers can use arbitrary dimensions and have varying capacity. The mixers can themselves be Dimension Mixer Model.
    Achieving dense parameterization is possible when mixing is performed such that there is a path from any input to any output dimension. 
    }
    \label{fig:dimension_mixer_general}
  \end{center}
  \vspace{-10pt}
  \vspace{1pt}
\end{figure} 

\IEEEPARstart{T}{he} recent success of various Neural Network Architectures such as Convolutional Neural Network (CNN)\cite{lecun2015lenet, krizhevsky2012imagenet, simonyan2014very, he2016deep}, Transformers\cite{vaswani2017attention, dosovitskiy2020image}, and MLP-Mixer\cite{tolstikhin2021mlp} can be credited to the sparse structured processing of input signals and the function or parameter sharing used in these architectures. 
These architectures are suitable for many downstream tasks, such as image classification, while working differently from each other. 
Although they share common parts, such as processing in patches, and sharing processing functions across tokens, they also mix signals at different locations in a variety of ways depending on the architecture. 

This abundance of processing the same input in multiple ways towards the same objective allows us to analyse and find common mechanisms among these architectures as well as differences as shown in Table~\ref{tab:temp_archi_comparision}. Moreover, these architectures are quadratic with respect to some of the input dimensions. For example, with sequence length in Attention, with channel dimension in Convolution or with patch/channel dimension in MLP-Mixer. We search for a general sparse signal processing mechanism allowing for sparsifying structured as well as unstructured dimensions of input tensor.

Partial signal processing is a recurring pattern of many deep architectures. 
Architectures such as Coupling-Flows\cite{dinh2014nice, gomez2017reversible} suggest that partial mixing of inputs is sufficient for dense unstructured input signals and provides the benefit of invertibility. 
ShuffleNet\cite{zhang2018shufflenet, ma2018shufflenet} shows that efficiency can be increased with decreased parameters if we use partial channel mixing along with channel permutations. 
These architectures motivate us to search for a general signal processing method with good scaling properties, capable of processing inputs sparsely in a series of stages.

Processing structured data with neural networks relies on model's inductive biases, such as translation eqivariance with 2D images, leading to highly parallelizable and parameter efficient processing.
However, for unstructured input dimensions, general dense networks do not scale well with the dimensionality of the input signal, and thus the default fully-connected models present a scaling obstacle.
While sparse parameterizations can partially alleviate this issue, general sparse matrix multiplication algorithms are lagging behind their dense counterparts in terms of performance. 
On the other hand, low-rank matrix parameterizations reduce the complexity by applying associativity of dense low-rank components multiplication\cite{idelbayev2020low} and lead to better performance. 
Structured sparse matrices (butterfly matrices) such as seen in the Fast Fourier Transform (FFT)~\cite{cooley1965algorithm} and recently in neural networks~\cite{prabhu2020butterfly, chen2021pixelated, dao2022monarch} show great promise in parameterizing dense matrices and efficient matrix multiplication.

\begin{table*}
    \centering
    \caption{
    Overview of properties of various architectures.
    }
    \resizebox{\linewidth}{!}{
    \begin{tabular}{@{}lccccc@{}}

\toprule

Model & Unit & Strictured & Input Sparsity & Mixing & Dense Operation \\
\midrule
Linear (or LowRank) & dot-product & \xmark & None & 1-layer & - \\
Linear-Butterfly & linear & \xmark & block sparse & butterfly-factors & block-linear  \\
MLP-Butterfly (ours) & MLP & \xmark & block sparse & butterfly-factors & block-MLP  \\
\midrule
CNN & convolution & \cmark & sliding window & radial & conv channels \\
% CNN-Mixers (ours) & group-conv & \cmark & sliding window & radial\&but-fact* & block-conv-channels \\
MLP-Mixer & MLPs & \cmark & patch \& channels & 1-block & channel+patch MLP  \\
Patch Only MLP (ours) & MLPs  & \cmark & patch & radial & patch MLP \\
\midrule
Transformer & Attention \& MLP  & \cmark & tokens \& sequence & 1 block & KQV, Attn \& MLP  \\
Butterfly Attn. Tranf. (ours) & Attention \& MLP  & \cmark & tokens \& sequence & butterfly-factors & KQV \& MLP  \\
% Parallel Token Attn. (ours) & KQV \& MLP  & \cmark & tokens \& sequence & 1-block & Attn \& MLP  \\
\midrule
Dimension Mixer & Any complete mix & Any & Group $\leq$ & Mix all dims & Any \\
(Generalization) & $A \to B$ transform &  & Total Dims & in $L$ layers &  \\

\bottomrule
\end{tabular}
    }
    \label{tab:temp_archi_comparision}
\end{table*}

Armed with these observations, 
we propose a general sparse signal processing model termed the Dimension Mixer model. 
It leverages group-wise mixing of input signals, processed in layer-wise fashion such that they communicate with each other, i.e. for $\mathbb{R}^N \to \mathbb{R}^M$ transformation, all $N$ input dimensions can have non-zero derivative with respect to all $M$ output dimensions. 
In other words, there must exist a mixing path from each input dimension to each output dimension. 
Dimension Mixer consists of several parallel dimension \emph{Select} and \emph{Mix} stages, as demonstrated in Fig.~\ref{fig:dimension_mixer_general}. 
This mechanism is repeated in a sequence of layers such that input dimensions can be mixed densely. 

Inspired by the widely used sparse signal mixing method from the FFT, which uses butterfly matrices and block-wise processing, we generalize the butterfly structure beyond Linear Transform to Non-Linear mixing function. 
Since such mixing borrows efficiency and scalability from FFT, it has $\mathcal{O}(N \log_r N)$ complexity, where $N$ represents the number of dimensions and $r$ represents mixing block size (``Radix-$r$''). 

Although butterfly structures have been previously explored in both learned and non-learned settings to replace dense matrix multiplication, our primary contribution is the extension of this paradigm to non-linear setting while retaining the structured mixing of different dimensions of the input tensor. 

We propose an efficient Butterfly Attention Mixer, an approximation of the self-attention mechanism, with sub-quadratic complexity in Sequence Length(S).

We additionally devise a new mixing strategy for 2D images called Patch-Only MLP Mixer. It lies between the original MLP Mixer and a CNN; this view helps to unify the working mechanism of both architectures. 
To process 2D images, we propose use of only patch-level mixing strategy.

We conduct studies of varying scale to evaluate efficiency and approximation capability our methods in several datasets.

Experiments on use of sparse MLPs in MLP-Mixer architecture on CIFAR-10 classification dataset shows that Butterfly MLP can sparsify the dense MLP dimension of MLP-Mixer producing better results than Sparse Linear. 

Furthermore, we find that sparse Butterfly Attention is sufficient for CIFAR-10 and CIFAR-100 datasets, produces faster architectures and scales better for large sequence lengths as well. We largely credit the performance to structured nature of butterfly attention when used in image.
Experiments on the Long Range Arena (LRA) Benchmark also demonstrate that the Butterfly Attention performs better than the baselines on the Retrieval and Image tasks with best Average accuracy.

Finally, we demonstrate the ability of our method to solve the challenging Pathfinder-X task, with the sequence length of 16,384 tokens. 
This task remained untackled by many low-rank attention methods; this attests to the generality and efficiency of our method.

The experiments on Patch-Only MLP-Mixer shows that our architecture produces smaller hidden representation and allows for more computational efficient mixing of image signal.
\section{Background and Related Works}
\label{sec:background}

\textbf{Structured Processing in Deep Learning}
One of the main ingredients to the success of deep learning frameworks is using structured processing units to handle structured data. These processing units enable the parallel processing of structured data, leveraging the power of accelerated computing technology such as GPUs and TPUs. The Convolutional Neural Netoworks~\cite{lecun2015lenet,krizhevsky2012imagenet,he2016deep}, in general, have sliding window filters which apply a linear transformation to patches of the image. From the perspective of signal mixing, the CNN architecture generally works by increasing the receptive field of filters as the number of layers increases. This allows the later layers to attend to a larger region of the input image, although indirectly through previous kernels. The CNN was the only dominant architecture for vision till the advent of Vision Transformers (ViT)~\cite{dosovitskiy2020image}. ViT generally works with non-overlapping patches, and the mixing of the patches happens immediately by the attention layer without waiting for later layers to comprehend the whole input signal. It is followed by the processing the signals per patch/token by the MLP layer. However, Vision Transformers lack a sliding window, thus preventing the shift equivariance inductive bias of CNNs. Contrary to this, MLP-Mixer~\cite{tolstikhin2021mlp}
replaces an attention layer with a channel mixing layer which is equally effective. 

ViT~\cite{dosovitskiy2020image} scales poorly against large sequence lengths, thus preventing the division of images into  a large number of patches. Later, Swin Transformer~\cite{liu2021swin} RegionViT\cite{chen2021regionvit} tackled such problems using convolutional priors and hierarchical attention. Majority of the works in transformers for efficiency~\cite{tay2020efficient} focus on creating sparse attention pattern~\cite{tu2022maxvit, hassani2022neighborhood, chen2021scatterbrain, chen2022vit, beltagy2020longformer, child2019generating, correia2019adaptively, xiong2021nystromformer, kitaev2020reformer, wang2020linformer, choromanski2020rethinking, tay2021synthesizer, wang2020cluster}. While some of the works also focus on making the sparse MLP block~\cite{fedus2021switch}.

Recently, the volume of literature in MLP-Mixer is also getting equally bigger\cite{liu2022we, tang2022image, chen2021cyclemlp, wang2022dynamixer, guo2022hire, zheng2022mixing, yu2021s, hou2022vision}. Most of these studies investigate the different 
ways of processing the input signals such as gating-based mixing~\cite{liu2021pay}, and shift in channels~\cite{lian2021mlp}.
In summary, the majority of the works in this category involve mixing either the patches and/or channels in various ways. 
These works motivated us to design a Patch-Only MLP-Mixer, which only mixes patch-wise for signal processing similar to CNN.

\textbf{Partial Signal Processing}
ShuffleNet~\cite{zhang2018shufflenet} uses group convolution over the channel dimensions, allowing for efficient convolution due to the reduced number of channels. It also mixes the channels for evenly distributing the output signals to the next convolution layer. It is important to note that AlexNet\cite{krizhevsky2012imagenet} uses parallel grouped convolution for accelerating on two GPUs and combines the hidden states in some layers to process them jointly. Similarly,  Megatron-LM\cite{shoeybi2019megatron} splits the input tokens into two blocks, process them independently and again combines them at the end of the attention and MLP blocks. Such a \emph{split}, \emph{process}, and \emph{combine} method allows for processing a large number of tokens in a parallel and efficient manner.

Coupling Flows~\cite{dinh2014nice, kingma2018glow, jacobsen2018revnet, ho2019flow++} and Reversible ResNet~\cite{gomez2017reversible}, use split and process mechanism that enables invertibility. ~\cite{teshima2020coupling} shows partial mixing of signals 
can approximate any diffeomorphic function. Furthermore, partial processing makes the computation of the Jacobian efficient. The key takeaway with these architectures is that partial signal processing is sufficient for function approximation and provides efficiency and scalability benefits. These works motivate our work Dimension Mixer model.

\textbf{Sparse Linear and Non-Linear Models}
The complexity of matrix multiplication for a vector input is known to be $N^2$, where $N$ is the input dimension. This problem has been tackled to some extent using Fast Matrix Multiplications~\cite{fawzi2022discovering, strassen1969gaussian} and Low-Rank Matrix Decomposition. The low-rank transformation has been widely used in efficient CNN architectures~\cite{peng2017large, liu2022more}. Models like EffecientNet-v2~\cite{tan2021efficientnetv2} and  MobileNets~\cite{howard2017mobilenets} use depth-wise and point-wise convolution as a low-rank factorization of a standard convolution.

One can get a highly sparse matrix using pruning-based techniques~\cite{han2015learning}. However, these methods accelerate neural networks on CPUs and mobile devices but fail to accelerate significantly on GPUs. 

\textbf{Linear Butterfly Sparsity}
The block sparse matrices~\cite{gale2020sparse} have been used widely to accelerate matrix multiplication. Another alternative is to use butterfly matrices~\cite{prabhu2020butterfly, dao2019learning}, which is inspired by the FFT. These matrices scale well with dimensions as well. Some of the past works~\cite{prabhu2020butterfly, chen2021pixelated, dao2022monarch, fan2022adaptable} have successfully used linear butterfly transformation to replace dense transformation and have produced highly efficient architectures.
In this paper, we generalize the Butterfly Sparsity to arbitrary non-linear dense operation. Furthermore, structure of butterfly sparsity is block sparse, thus highly parallelizable and have highly efficient hardware~\cite{fan2022adaptable} and software implementations.
The sparsity is simple, uses commutation, allows varying sparsity and does not require masks.

\textbf{Long Range Arena (LRA)}
is one of the most challenging benchmarks for evaluating the quality of a model in long-range sequences~\cite{tay2020long}.
The benchmark consists of several unique sequence processing tasks with varying sequence length of more than one thousand tokens. 
ListOps~\cite{nangia2018listops} consists of hierarchical mathematical operations which measure the parsing and analytical abilities. 
Byte-level text classification~\cite{maas2011learning} is a binary classification task on characters, requiring complete access to all tokens. 
Byte-level document retrieval~\cite{radev2013acl} measures the ability to create a compressed representation of input; it counts as a binary classification problem. 
Similarly, Image Classification Task~\cite{krizhevsky2009learning} defined on a sequence of flattened pixels measures the ability to learn 2D relation between tokens without explicit 2D inductive bias. 
Pathfinder task, including the Pathfinder-X~\cite{linsley2018learning} is a synthetic dataset of labyrinths for measuring learning and inference abilities on long-range spatial dependencies.
Succeeding on all of these tasks shows that an efficient Attention Mechanism works on diverse and challenging tasks formatted as long sequence problems. 
The dataset provides a medium for fair comparison using similar hyperparameters, rather than a competition using different configurations.

\section{Dimension Mixer Models}

Dimension Mixer model is a simple and general structure observed among most deep learning architectures. The key observation is that of structured signal processing and performing mixing of all input signals efficiently. 
Computational efficiency is achieved thanks to partial (group-wise) mixing of input signals. For structured data, the mixing is itself structured similarly to the data and mixing functions like MLPs can be shared. Parameter efficiency is gained due to function sharing as well as partial mixing.

Geometric Deep Learning~\cite{bronstein2017geometric, bronstein2021geometric} tackles the generality of these architectures by considering graphs as an underlying data structure, however, we generalize from the perspective of signal mixing and processing. Dimension Mixer model allows us to develop structured processing for unstructured data as well. 
Fig.~\ref{fig:dimension_mixer_general} shows a simple example of a general Dimension Mixer model using two major operations (i) Select a group of dimensions and (ii) Process or Mix the selection. This operation is carried out in parallel as well as sequentially for efficient mixing of signals. The signal mixing can be evaluated with Graph Theory for the effective flow of signals.

The structure of Dimension Mixer model can be arbitrary in terms of input grouping/selection and mixing/processing function, as long as the input signals are efficiently mixed. The butterfly structure of the FFT is highly efficient, especially for GPUs. Hence, we focus on using the Butterfly Structure for its immediate utility to accelerate current Neural Network Architectures.

\subsection{Non-Linear Butterfly Mixer}

\begin{figure*}
    \centering

    \begin{subfigure}{0.65\textwidth}
    \includegraphics[width=0.99\linewidth]{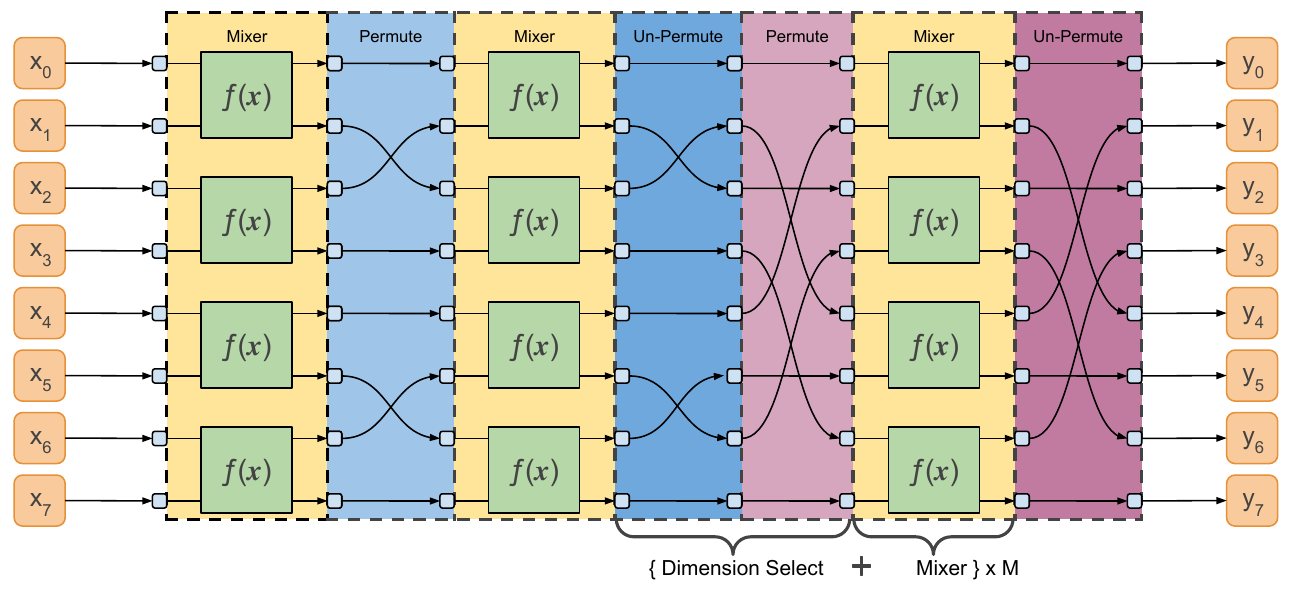} 
    \end{subfigure}
    \begin{subfigure}{0.3\textwidth}
    \includegraphics[width=0.99\linewidth]{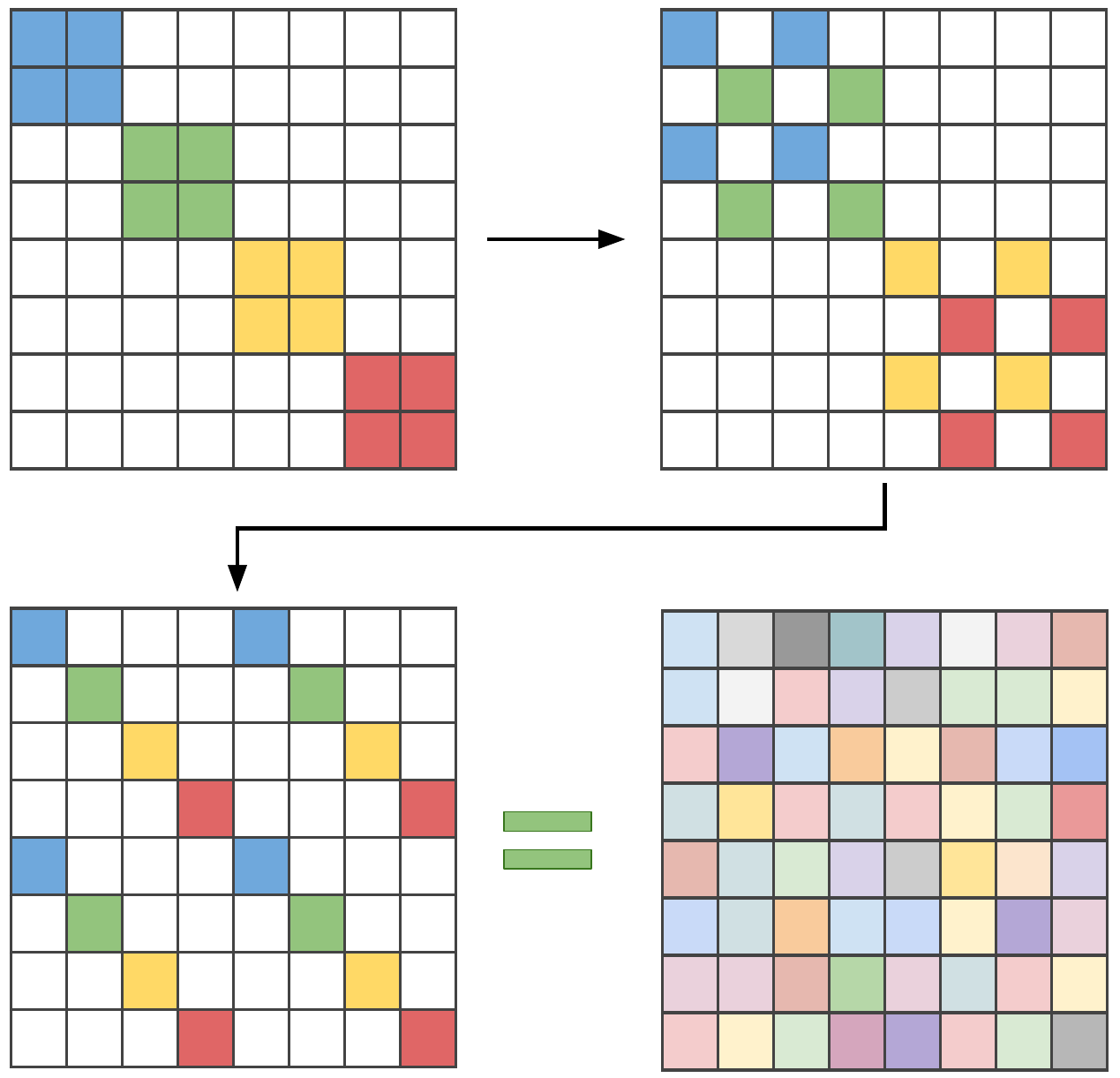}
    \end{subfigure}
    
    \caption{An example of FFT style Non-Linear Butterfly Mixer. This example shows the mixing of an 8-dimensional input signal using Radix-2 Butterfly. The first layer selects the dimension as it is. However, later layers use Permute to bring different dimensions in a block and later perform un-permute to place the dimension in their original location. For Radix-4 Butterfly a mixer block takes 4 dimensions as input and permutes accordingly as shown in Algorithm~\ref{algo:butterfly_mlp}.}
    \label{fig:butterfly_dimension_mixer_model}
\end{figure*}

Although butterfly structure has been used widely in Linear Models~\cite{prabhu2020butterfly, dao2022monarch, chen2021pixelated}, our theory of Dimension Mixer model allows non-linear mixing for effective mixing of input signals. We propose a non-linear butterfly structure model as shown in Fig.~\ref{fig:butterfly_dimension_mixer_model}. The linear butterfly transformation is a special case with linear mixing function $f(x)$. Furthermore, we may use any learnable function as a mixing function and use butterfly structure to sparsify any dimension of input signal. This generalization allows us to extend butterfly structure beyond linear transforms.
We can use MLP itself as non-linear mixing function to mix unstructured input dimension or Transformer as mixing function to mix long sequences. 

To this end, we propose unstructured Non-Linear Mixers called Butterfly MLP and Butterfly Attention. 

\begin{algorithm*}
\caption{Implementation of Block-Sparse and Butterfly MLP}
\label{algo:butterfly_mlp}
\begin{minted}[fontsize=\footnotesize,baselinestretch=1]{python}
class BlockLinear:
    def __init__(self, num_blocks, input_block_dim, output_block_dim):
        self.weight = torch.randn(num_blocks, input_block_dim, output_block_dim)
        self.bias = torch.randn(num_blocks, 1, output_block_dim)
        
    def forward(self, x):
        ## x -> [num_blocks, batch_size, input_block_dim]
        return torch.batch_matmul(x, self.weight) + self.bias
    
class BlockMLP:
    def __init__(self, input_dim, layer_block_dims=[], actf=nn.GELU):
        self.block_dim = layer_dims[0]
        num_blocks = input_dim//layer_block_dims[0]
        
        ### Create a block MLP
        self.mlp = nn.Sequential([])
        for i in range(len(layer_block_dims)-1):
            self.mlp +=[ BlockLinear(num_blocks, layer_block_dims[i], layer_block_dims[i+1]),
                         actf() ]
        self.mlp = self.mlp[:-1]
        
    def forward(self, x):
        bs, input_dim = x.shape
        x = x.view(bs, -1, self.block_dim).transpose(0,1)
        x = self.mlp(x)
        x = x.transpose(1,0).view(bs, -1)
        return x
    
## x : Input with shape [batch_size, input_dim].
## block_dim -> size of block in block_sparse MLP. Usually a factor of input_dim
## block_layers : Layers of block mixing function.
## fn_block : Block mixing function; is parallel non-linear mixer operating per block.
## y : Output with same shape as Input x for simplicity.

block_layers = []
for _ in range(log_base(input_dim, base=block_dim)): ## using hidden expansion of 2
    block_layers += [ BlockMLP(input_dim, [block_dim, block_dim*2, block_dim]) ]

## Using Butterfly Permutation
for i, fn_block in enumerate(block_layers):
    stride = block_size**i if ( block_size**(i+1) <= input_dim ) else input_dim//block_size
    x = x.view(-1, block_dim, stride).transpose(2, 1).view(batch_size, -1)
    x = fn_block(x)
    x = x.view(-1, stride, block_dim).transpose(2, 1).view(batch_size, -1)
return x
\end{minted}

% # x : Input with shape [batch_size, input_dim].

\end{algorithm*}

\subsubsection{Butterfly MLP} 
Butterfly MLP uses MLP as a mixing function. 
We implement a column of MLPs in parallel which helps utilize the acceleration of GPUs. 
This simply breaks a whole MLP into blocks of MLPs in a butterfly structure for efficient mixing. 
Thus produced architecture is parameter efficient as well as allows having arbitrary MLP design (in terms of depth, width and activation) as a mixing function. 
Our method is more flexible than using Butterfly Linear as a replacement for a dense layer. 
If we take the Radix-$N$ butterfly for $N$-dimensional input, it becomes a single mixing using the dense method. 
Hence, Butterfly MLP generalizes to standard MLP. 
However, Radix-$M$ ($M<N$) butterfly structure shards MLP into multiple smaller MLPs which is more parameter efficient, as well as highly parallelizable. The pseudocode for Butterfly MLP is shown in Algorithm~\ref{algo:butterfly_mlp}    

% \textbf{2D Grid Bilinear Mixer} We use the 2D grid for approximating the mixing function. We use an evenly spaced grid and learn the $Y$ values. We can simply use the input to index the neighbourhood points and apply bi-linear interpolation. This model allows us to use 2D grid-based activation function on MLPs. We implement this structure on the PyTorch framework, however, due to inefficiency, we implement the CUDA kernel which is highly efficient and scales well with grid size. This method consumes high memory if the input dimension is high or if the grid size is large. Our implementation has more areas to improve upon, however, the current implementation shows that it can work successfully.

% This model also demonstrates the generality of the Dimension Mixer model beyond MLPs. 

% This method allows us to benefit from the property of the grid in that it has a constant computational complexity for the size of the grid. 

\subsubsection{Butterfly Attention}

\begin{figure*}
    \centering
    \begin{subfigure}{0.47\textwidth}
    \includegraphics[width=0.99\linewidth]{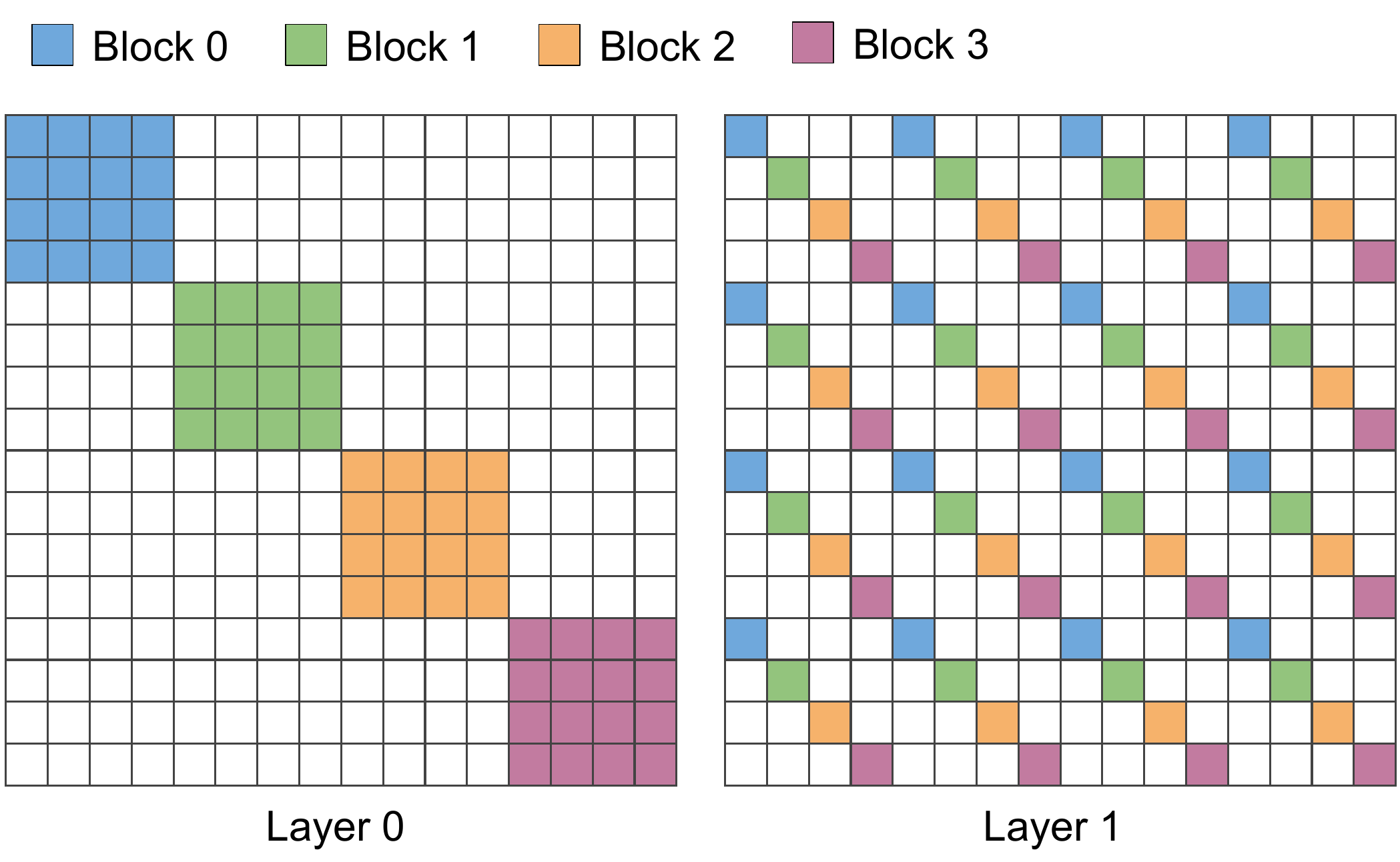} 
    \caption{Butterfly Attention}
    \label{fig:butterfly_attention}
    \end{subfigure}
    \begin{subfigure}{0.52\textwidth}
    \includegraphics[width=0.99\linewidth]{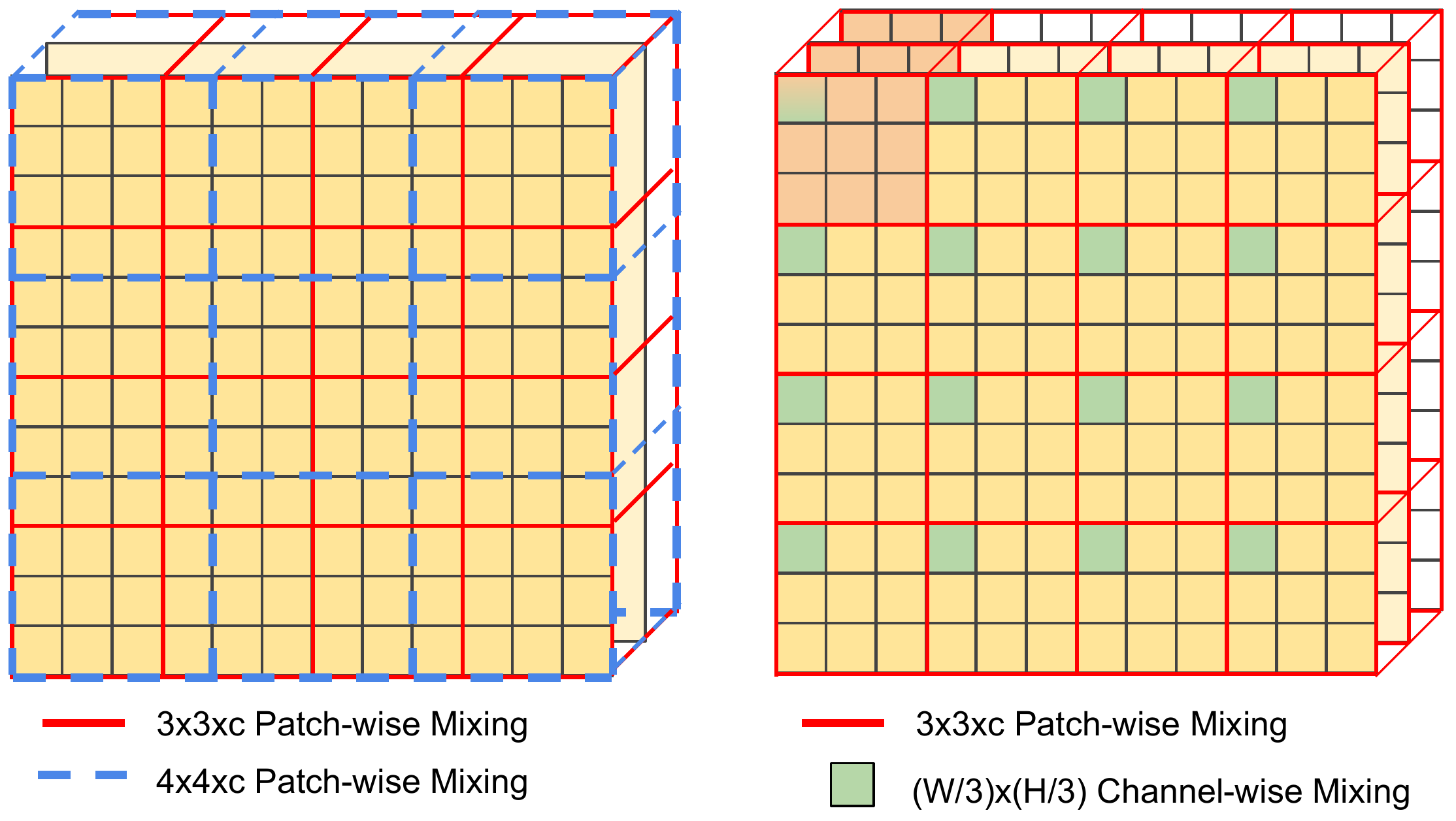}
    \caption{MLP-Mixer Comparison}
    \label{fig:patch_only_mlp_mixer}
    \end{subfigure}
    
    \caption{\textbf{(a)} An example of Butterfly Attention pattern on sequence length of 16 with butterfly structure of Radix-4. Using Radix-$\sqrt{N}$ creates two sparse attention matrix for complete mixing of signals. \textbf{(b)} \textit{(left)} Patch Only MLP mixer (ours) compared to \textit{(right)} Patch-Wise and Channel-Wise Mixing for 12x12 image size. The Channel-wise Mixing is replaced by Different size Patch-wise Mixing by our method.}
    \label{fig:att_mlpmixer}
\end{figure*}

We also apply the efficient mixing of butterfly structures in the Attention mechanism of Transformers architecture (different from Pixelated Butterfly~\cite{chen2021pixelated} and \textit{ABfly}~\cite{fan2022adaptable}). It is widely known that the complexity of Attention is $S^2$ with sequence length $S$. Furthermore, there is a large number of cases where the sequence length can blow up significantly such as on large images when using small patch size, on long documents and on audio/video signals. To solve this issue, we apply partial block-wise attention using the butterfly structure , which reduces the complexity of Attention to 
just $S$, however it would require $\log_a(S)$ layers of Attention for complete mixing of tokens where $a$ is Radix or block size. For experiments, we use Radix-$\sqrt{S}$ for mixing which creates complete mixing within two attention blocks (see Fig.~\ref{fig:butterfly_attention}). 
Since the MLP layer is invariant to permutation of tokens, we apply butterfly structure to Transformer Block as a whole as shown by the Algorithm~\ref{algo:butterfly_attention}. Since the Butterfly Attention converts Blocks of Tokens into batches, we can use any implementation of transformer architecture, including the optimized ones like Flash Attention~\cite{dao2022flashattention}, or even Low-Rank Attention like~\cite{xiong2021nystromformer}.  

Despite having partial attention, experiments show that it performs better than full attention when used on CIFAR-10, CIFAR-100 datasets and on LRA benchmark. Furthermore, the experiments show that the Butterfly Attention mechanism is faster for training as well as memory efficient.

\begin{algorithm*}[h]
\caption{Permutation of Butterfly Attention for Transformers}
\label{algo:butterfly_attention}
\begin{minted}[fontsize=\footnotesize,baselinestretch=1]{python}
# x: Input with shape [BATCH SIZE (B), SEQUENCE LENGTH (S), MODEL DIMENSION (D)]
# mask : Attention mask (binary) - EITHER: token-wise mask 
# - OR: same size as Attention [-1, num_blocks, num_heads, block_size, block_size]
# block_size (a): Radix or Block size of Butterfly Attention
# i : Index of layer in butterfly (i.e 0, 1, .. log_a{S}-1 ; S is Sequence Length)
# transformer: a transformer layer with Attention and MLP layers [Vaswani et el.]

B, S, D = x.shape
for i, transformer in enumerate(transformers_layers):
    stride = block_size**i if ( block_size**(i+1) <= S ) else S//block_size
    
    x = x.view(B, -1, block_size, stride, D).transpose(2, 3).view(-1, block_size, D)
    x = transformer(x, mask)
    x = x.view(B, -1, stride, block_size, D).transpose(2, 3).view(B, S, D)

return x
\end{minted}

% stride = block_size**i
%     if stride*block_size > S:
%         stride = S/block_size
    
\end{algorithm*}

\subsection{Patch-Only MLP-Mixer for Vision}

In the MLP-Mixer architecture, the mixing between two or more patches is done by token mixing (or channel-wise mixing) MLP. However, in CNN architecture, the only use of patch-wise mixing is sufficient. This is due to the sliding window used in convolution which helps to mix the signals of different regions along with an increase in the receptive field with an increase in depth. We search for a mechanism that allows for mixing using patch-only but without the sliding window. 

We propose Patch-only MLP Mixer (Fig.~\ref{fig:patch_only_mlp_mixer}). The architecture consists of Image of size $I$ and Kernels/Patches of size $K_1$, $K_2$, $K_3$, $\cdots$ where, $I$ = $K_1 \times K_2 \times K_3 \cdots $ such that $K_i$ and $K_j$ do not have common factors for different $i$ and $j$. If patch-sizes have common factors then the mixing occurs in different partitions corresponding to the factors without complete mixing. For example, if we have patches of size 6 and 8, then mixing happens in $lcm(6, 8) = 24$ instead of 48 block size. We choose only 2 factors for patch size for the simplicity of experiments. Patch Only MLP-Mixer lies in between MLP-Mixer and Convolution as it uses MLP for processing patches but only uses patches for overall input mixing.
\section{Experiments}
We share the code \href{https://github.com/tsumansapkota/Dimension-Mixer}{\underline{A}}, \href{https://github.com/tsumansapkota/Solving-LRA-PathX}{\underline{B}}, \href{https://github.com/bhattarailab/pathfinder-generator}{\underline{C}} to reproduce the experiments.

\paragraph{Butterfly MLP}

We test the approximation capacity of (i) our Butterfly MLP, (ii) MLP with Butterfly-Linear Transform~\cite{lin2021deformable, dao2022monarch} and (iii) Dense MLP in a MLP-Mixer settings trained on CIFAR-10 dataset. We find that Butterfly-MLP performs better than Butterfly-Linear with slightly more parameters and MACs. Both sparse models have large savings on parameters and compute.  For the experiments, we use MLP-Mixer with 7 layers and train for 200 epochs with Adam optimizer(lr=0.001) and cosine decay lr. The dimension of patch and channel are 64 and 121 respectively - square numbers to create block size of 8 and 11 respectively.   
In Table~\ref{tab:butterfly_in_mixer}, we show results for comparison with hidden expansion of 1 and 2 on the MLP layer. Butterfly MLP has extra parameters consisting of biases in each mixing block of MLP unlike Butterfly Linear based MLP. The number of permutations (including the un-permute) used in MLP block shows that Butterfly-Linear requires 4 such permutations in 2 layers of sparse weights that follows MLP structure. However, Butterfly MLP accomplishes this with only 2 permutation and on a single butterfly structure. Here, a single block is itself non-linear and doesn't require two layers for approximating an MLP. Theoretically, this reduces data movement between parallel and sequential blocks.

\begin{table*}
    \centering
    \caption{
    Replacing Dense layer with Sparse layers in MLP-Mixer. The accuracy is calculated over 8 seeds.
    } % table caption is always above the table
    \resizebox{0.8\linewidth}{!}{
        \begin{tabular}{@{}rcccccc@{}}

\toprule 

Method & Expansion & Parameters & MACs  & Acc & Max-Acc & Permutations \\
% \cmidrule(lr){0-5}
\midrule

Dense & \multirow{3}*{1}& 351.68 k & 20.73 M & 83.83$\pm$0.407  & 84.64 & 0 \\
Butterfly-Linear & & 140.96 k & 5.0 M & 82.45$\pm$0.361  & 83.01 & 4 \\
Butterfly-MLP & & 143.55 k & 5.33 M & 82.68$\pm$0.244  & 83.06 & 2 \\
\midrule
Dense & \multirow{3}*{2}& 615.29 k & 40.89 M & 84.07$\pm$0.355  & 84.47 & 0 \\
Butterfly-Linear & & 193.86 k & 9.34 M & 83.49$\pm$0.322  & 83.84 & 4 \\
Butterfly-MLP & & 197.75 k & 9.88 M & 83.70$\pm$0.302  & 84.26 & 2 \\

\bottomrule

\end{tabular}
    }
    \label{tab:butterfly_in_mixer}
\end{table*}

\paragraph{Butterfly Attention}

\begin{table*}
    \centering
    \caption{
    An experiment comparing Butterfly Attention with Dense Attention. 
    The B-Attention column represents if the Butterfly Attention is used. 
    The Sequence Length for patch size 4, 2 and 1 are 64, 256 and 1024 respectively. Similarly, the embedding dimension for patch size 4, 2 and 1 are 128, 64 and 64 respectively. The device used for measuring time is GTX 3090 with PyTorch-2 and an off-the-shelf implementation of Transformers.
    } % table caption is always above the table
    \resizebox{\linewidth}{!}{
        \begin{tabular}{@{}rccccccccccccc@{}}

\toprule 

Dataset & \multicolumn{8}{c}{CIFAR-10} & \multicolumn{4}{c}{CIFAR-100} \\

\cmidrule(lr){2-9} \cmidrule(lr){10-13}

Layers, Patch & \multicolumn{2}{c}{8, 4} & \multicolumn{2}{c}{4, 4} & \multicolumn{2}{c}{4, 2} & \multicolumn{2}{c}{4, 1} & \multicolumn{2}{c}{4, 4} & \multicolumn{2}{c}{4, 2} \\

\cmidrule(lr){2-3} \cmidrule(lr){4-5} \cmidrule(lr){6-7} \cmidrule(lr){8-9}  \cmidrule(lr){10-11} \cmidrule(lr){12-13}

B-Attention & No & Yes & No & Yes & No & Yes & No & Yes & No & Yes & No & Yes \\

\midrule

Accuracy ($\uparrow$) & 81.90  & 84.69 & 81.04 & 83.68 & 78.35 & 80.80 & 65.65 & 73.96 & 53.92 & 57.02 & 50.03 & 53.29 \\

Time (ms) & 17.41  & 17.32  & 9.34 & 9.45 & 16.88 & 9.53 & 198.04 & 26.72 & 9.56 & 9.64 & 32.53 & 15.61 \\

Memory (MiB) & 434  & 386 & 304 & 270 & 1300 & 418 & 16592 & 1698 & 434 & 380  & 2468 & 810 \\

\cmidrule(lr){2-3} \cmidrule(lr){4-5} \cmidrule(lr){6-7} \cmidrule(lr){8-9}  \cmidrule(lr){10-11} \cmidrule(lr){12-13}

Params. (M) & \multicolumn{2}{c}{1.148} & \multicolumn{2}{c}{0.618} & \multicolumn{2}{c}{0.299} & \multicolumn{2}{c}{0.79} & \multicolumn{2}{c}{1.355} & \multicolumn{2}{c}{1.773} \\

\bottomrule

\end{tabular}

    }
    \label{tab:sparse_attention}
\end{table*}

We test the capacity of Butterfly Attention as compared to dense Attention proposed in the original paper. We experiment on CIFAR-10 and CIFAR-100 datasets. We experiment on different patch sizes and a different number of layers without using Positional Encoding. In Table~\ref{tab:sparse_attention} we show the accuracy, training time, memory usage and parameters of the vision transformers. The time taken and the memory usage is given by average values of 50 steps of training. The experiments are designed to test the resource consumption and accuracy of both sparse and dense attention models. Experiments show that Butterfly Attention scales better with longer sequence lengths, and the accuracy is comparable to dense attention. Our method produces a faster, less memory consuming and better-performing model as compared to dense attention. We discuss the possible reasons for accuracy gains in Appendix~\ref{appendix:attention_in_vit}. The experiments are carried out for 300 epochs and 64 and 128 batch size for CIFAR-10 and CIFAR-100 respectively with cosine decay of learning rate 0.0001. We use a total of 8 Attention heads on all the models. 

\paragraph{Long Range Arena} 
We conduct experiments on the Long Range Arena (LRA)~\cite{tay2020long} benchmark to test the capacity of Butterfly Attention with regards to context length. 
The LRA benchmark is designed for a fair comparision of effecient attention mechanisms using the same settings: small lightweight networks, and generalization on a diverse range of tasks. 
Until recently, with these constraints, the most complex task -- the Pathfinder-X with 16K tokens sequence length -- was not solved. 
This work was among the first ones to ignite interest to very long sequence length modeling, and many Transformer papers started using LRA and Pathfinder-X as a \textit{model} performance benchmark instead of arena for \textit{attention}.

The data for other sparse attention methods and standard attention is from NystromFormer~\cite{xiong2021nystromformer}. 
We use the same training and evaluation protocol for our methods as well. 
The experiments on LRA (except Pathfinder-X) consist of a simple Xformer architecture with 2 layers and 2 attention heads.
We use the learning rate of 0.0001 with warmup, and linear learning rate decay with 0.1 dropout for embedding, attention, MLPs, and residuals (refer to ~\cite{xiong2021nystromformer} for all hyperparameters). 
The results on LRA along with their average accuracies are shown in Table~\ref{tab:butterfly_attn_lra}.
The average does not include the Pathfinder-X result, as we use different architecture $\to$ 4 layers and 4 attention heads in the network. We also modify dropout location, add weight-decay and use different training hyperparameters. The exact method and steps we used for solving Pathfinder-X is explained in Appendix ~\ref{appendix:solving_pathx}. Experiments show that our method performs competitive among tasks with best average score.

\begin{table*}
    \centering
    \caption{
    Comparison of attention mechanisms on the Long Range Arena benchmark. 
    The Standard attention~\cite{vaswani2017attention} is a reference performance baseline. 
    Among the low-rank attention methods, Butterfly scores the best on most LRA tasks including the challenging PathFinder-X. 
    Token length of a task is indicated in parentheses.
    \textit{Legend: Flash-Attn~\cite{dao2022flashattention}, Reformer~\cite{kitaev2020reformer}, Linformer~\cite{wang2020linformer},  Performer~\cite{choromanski2020rethinking}, Nystromfrm.~\cite{xiong2021nystromformer}}
    }
    \resizebox{\linewidth}{!}{
        \begin{tabular}{@{}lcccccc|c@{}}

\toprule

Model & ListOps (2K) & Text (4K) & Retrieval (4K) & Image (1K) & Pathfinder (1K) & Average ($\le$4K) & Path-X (16K)\\ % Path-64 (4096) \\ % & Path-X (16384)\\

\midrule
\textit{Standard} & 37.10 & 65.02 & 79.35 & 38.20 & 74.16 & 58.77 & --- \\
\textit{Flash-Attn.} & 37.6 & 63.9 & 81.4 & 43.5 & 72.7 & 59.8 & 61.4 \\
\midrule
Reformer & 19.05 & 64.88 & 78.64 & 43.29 & 69.36 & 55.04 & --- \\
Linformer & \textbf{37.25} & 55.91 & 79.37 & 37.84 & 67.60 & 55.59 & --- \\
Performer & 18.80 & 63.81 & 78.62 & 37.07 & 69.87 & 53.63 & ---  \\
Nystromfrm. & 37.15 & \textbf{65.52} & 79.56 & 41.58 & 70.94 & 58.95 & ---  \\
Butterfly & 37.05 & 65.25 & \textbf{81.32} & \textbf{44.02} & \textbf{71.12} & \textbf{59.75} & \textbf{76.72}  \\ 

\bottomrule

\end{tabular}

    }
    \label{tab:butterfly_attn_lra}
\end{table*}

\begin{table*}[ht!]
    \centering
    \caption{
    Vision MLP Mixers Comparision. 
    We measure the Accuracy Parameters and MACs. MLP-Dims shows the dimension of two MLPs (m1 and m2) used in single block of mixing. The reported accuracy is best over 3 runs.}
    \vspace{1em}
    \resizebox{0.8\linewidth}{!}{
        \begin{tabular}{@{}rcccccccc@{}}

\toprule 

\multirow{2}{*}{Architecture} & \multirow{2}{*}{Layers} & MLP-Dims & \multicolumn{3}{c}{CIFAR-10} & \multicolumn{3}{c}{CIFAR-100}  \\

 \cmidrule(lr){3-3} \cmidrule(lr){4-6} \cmidrule(lr){7-9}
 &  & m1,m2 & Acc & Params & MACs & Acc & Params & MACs \\

 \midrule

MLP Mixer (c1) & \multirow{3}{*}{7}  & 81, 144 & 83.81 & 0.90M & 74.65M & 57.37 & 1.95M & 75.7M \\
Patch Only &  & 75, 147 & 84.66 & 0.81M & 23.04M & 55.55 & 1.14M & 23.37M \\
MLP Mixer (c2) &  & 64, 153 & 84.16 & 0.88M & 60.48M & 58.38 & 1.77M & 61.36M \\
\midrule
MLP Mixer (c1) & \multirow{3}{*}{10}  & 81, 144 & 83.03 & 1.23M & 106.36M & 56.34 & 2.28M & 107.41M \\
Patch Only &  & 75, 147 & 85.49 & 1.14M & 32.9M & 56.29 & 1.47M & 33.23M \\
MLP Mixer (c2) &  & 64, 153 & 84.20 & 1.22M & 86.16M & 57.81 & 2.10M & 87.04M  \\

\bottomrule

\end{tabular}

    }
    \label{tab:mlp_mixers}
\end{table*}

\paragraph{Vision MLP Mixers}

We compare our Patch-Only MLP mixer with the original MLP-Mixer architecture with a similar number of layers and a similar number of parameters in the mixing blocks. We try to balance the mlp dimension on both Mixer methods. However, due to different ways of scaling the input, we do not have same parameters.  
The experiments show that our method produces comparative or even better results than the original MLP mixer on CIFAR-10 and CIFAR-100 datasets (32x32 images). 

For MLP Mixer config-1 \textit{(c1)} we scale the input image to 36x36 size, extract patch of size 4 and expand channels by a factor of 3.
Furthermore, for config-2 \textit{(c2)} we use a patch size of 4 and expand channels by a factor of 3.2.
On Patch-Only MLP Mixer, we expand the image to 35 = 5x7 size and mix over the patch of sizes 5 and 7 with no channel expansion. We train all models for 200 epochs with Adam optimizer(lr=0.001) and cosine decay lr. We use 64 batch size for CIFAR-10 and 128 for CIFAR-100.

We compare two methods with a similar number of parameters in the mixing blocks. 
The results in Tab.~\ref{tab:mlp_mixers} show that our method performs competitively or even better than the original method. 
If we compare the speed of training these models, our method lags behind in wallclock time due to the Unfold and Fold operations needed to extract and combine patches.
However, these methods do not count towards MACs. 
In our configuration, MLP-Mixer produces higher MACs than our method. This is because MLP-Mixer produces large hidden image size, while our method produces smaller hidden image size and has a smaller classifier layer. 
Moreover, sparsity can be varied by selection of \textit{patch size} and \textit{hidden expansion}~\cite{hayase2023mlp}.

\section{Conclusion}
In this paper, we introduce a generic method of efficient input signal processing model called Dimension Mixer model.
We employed our method on multiple host architectures, such as MultiLayered Perceptron~(MLP) and Attention Layers of the Transformer Architecture, thereby introducing sparsity; to be exact, butterfly sparsity. 
A yet another contribution we made in this paper is the introduction of Patch-Only MLP mixer as an intermediate architecture between the original MLP-Mixer and the Convolutional Neural Network. 

All the proposed models are the application of the Dimension Mixer model which we find is insightful for analysing the signal processing on deep learning architectures and also for designing newer models. Experimental results show that our methods are often more efficient and/or more accurate than the counterpart architectures.

\textit{Limitations:}
Our study mostly uses small scale datasets for comparison. Benchmarking with existing models on large scale datasets like ImageNet~\cite{deng2009imagenet} can make our findings more significant.

\section*{Acknowledgments}

We thank Dr. Anton Obukhov for his help with running Pathfinder-X experiments on the Euler cluster of ETH Zürich.

{\appendices
\begin{table*}
    \caption{Summary of Pathfinder datasets used to solve pathfinder-X \textit{(Top to Bottom)}. Accuracy is reported per Methods for Butterfly Attention based Transformer. Here, some of the \textit{Pathfinder-32} and \textit{Pathfinder-64} samples are scaled for reference.}
    \resizebox{1.0\linewidth}{!}{
        \begin{tabular}{@{}cccccclc@{}}
\toprule
\rotatebox{90}{Image Size}&\rotatebox{90}{Contour Length}&\rotatebox{90}{Generated}&\rotatebox{90}{Num Distractors}&\rotatebox{90}{Paddle Gap}&\rotatebox{90}{Paddle Contrast}&Samples&Acc \\
\midrule
32&14&\xmark&-&-&-&\includegraphics[width=.75\textwidth,valign=M]{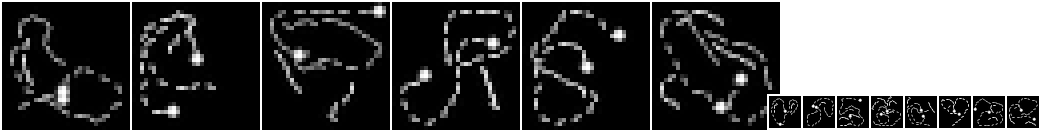}&69.17\\
64&9&\xmark&-&-&-&\includegraphics[width=.75\textwidth,valign=M]{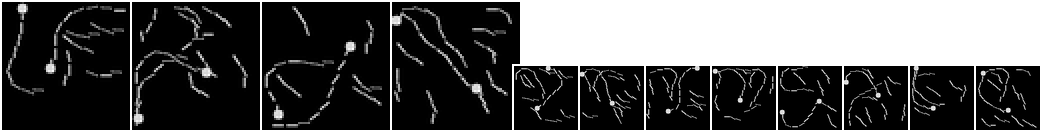}&80.99 \\
128&14&\cmark&5&1&0.9&\includegraphics[width=.75\textwidth,valign=M]{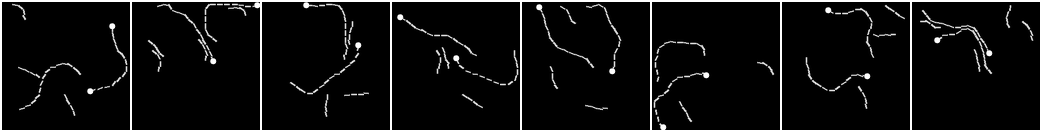}&71.84\\
128&14&\cmark&5&2&0.9& \includegraphics[width=.75\textwidth,valign=M]{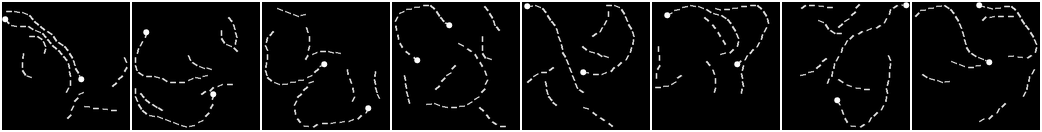}&75.05\\
128&14&\cmark&14&2,3&0.8&\includegraphics[width=.75\textwidth,valign=M]{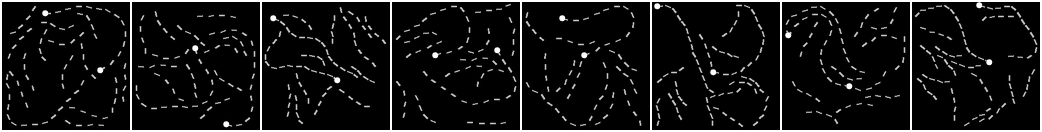}&75.46 \\
128&14&\cmark&20&2,3&0.73&\includegraphics[width=.75\textwidth,valign=c]{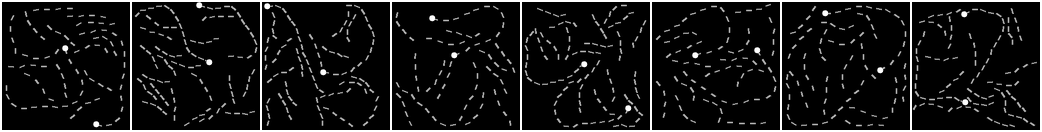}&76.41 \\
128&14&\xmark&-&-&-&\includegraphics[width=.75\textwidth,valign=c]{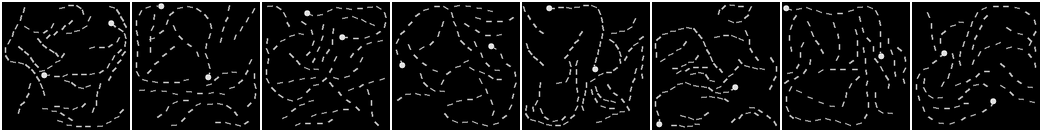}&76.72 \\
\bottomrule
\end{tabular}
    }
    \label{tab:solving_pathx}
\end{table*}

\section{Solving Pathfinder-X}
\label{appendix:solving_pathx}

Since brute force approach to training Pathfinder-X (\textit{Pathfinder128 contour-length-14}) did not work. We devised a step by step process to help the transformer grok the intermediate complexities. We follow same upscaling method from ~\cite{dao2022flashattention}, i.e. we use nearest neighbour interpolation of positional embedding and also increase the size of butterfly-block as well.

Experimentally, we find it easier to grok \textit{Pathfinder64 contour-length-9} than \textit{Pathfinder64 contour-length-14} initialized from \textit{Pathfinder32 contour-length-14}. To scale from \textit{Pathfinder64 contour-length-9} to \textit{Pathfinder128 contour-length-14} is a huge increase in complexity and makes it difficult to grok - it has many ~\textit{distractors}, have varying \textit{paddles/dash gap} and also seem to have lower \textit{contrast}. We generate datasets with increasing complexity to the equivalent of \textit{Pathfinder128 contour-length-14}.

Table~\ref{tab:solving_pathx} summarizes the datasets used and accuracy achieved by our method. The experiment use 4 layers of butterfly transformer to allow 2 complete mixing of signals using block size of $\sqrt{S}$, here $S$ is sequence length.

\begin{table}[h]
    \centering
    \caption{
    An experiment comparing Accuracy of ViT on CIFAR dataset with Structured Butterfly Attention, Randomization of Tokens on Butterfly Attention and Dense Attention. The Sequence Length for patch size 4, 2 and 1 are 64, 256 and 1024 respectively.
    }
    \resizebox{\linewidth}{!}{
        \begin{tabular}{@{}rccccccc@{}}

\toprule 

Dataset & \multicolumn{4}{c}{CIFAR-10} & \multicolumn{2}{c}{CIFAR-100} \\

\cmidrule(lr){2-5} \cmidrule(lr){6-7}

Layers, Patch & 8, 4 & 4, 4 & 4, 2 & 4, 1 & 4, 4 & 4, 2 \\

% \cmidrule(lr){2-3} \cmidrule(lr){4-5} \cmidrule(lr){6-7} \cmidrule(lr){8-9}  \cmidrule(lr){10-11} \cmidrule(lr){12-13}

% B-Attention & No & Yes & No & Yes & No & Yes & No & Yes & No & Yes & No & Yes \\

\midrule

Butterfly & 84.69 & 83.68 & 80.80 & 73.96 & 57.02 & 53.29 \\
Rand Butt. & 82.41 & 81.86 & 77.41 & 70.65 & 54.96 & 49.31 \\
Dense & 81.90  & 81.04 & 78.35 & 65.65 & 53.92 & 50.03 \\

\bottomrule

\end{tabular}

    }
    \label{tab:sparse_attention_random}
\end{table}

\begin{figure*}[h]
    \centering
    \includegraphics[width=1.0\linewidth]{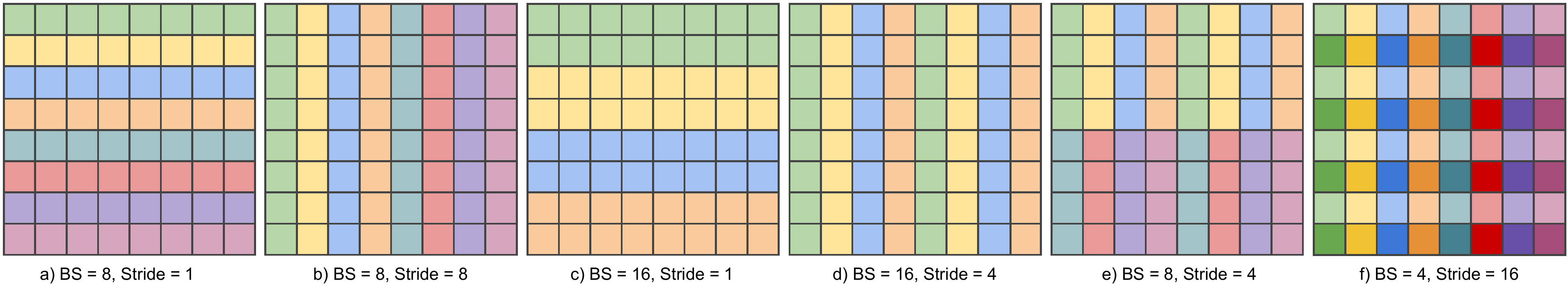} 
    
    \caption{
    The images with 8x8 = 64 tokens using butterfly attention of different block size (BS) and various strides. Each different color represents different blocks of attention. 
    The partial mixing of signals on multiple layers can create complete mixing of every tokens. Here, mixing combinations of $(ab)$, $(cd)$ $(cb)$, and $(cf)$ create complete mixing however, $(ae)$ and any same mixing like $(aa)$ does not mix every token.
    }
    \label{fig:butterfly_in_patch_image}
\end{figure*}

\section{Effect of Patches, Block Size and Stride in Butterfly ViT}
\label{appendix:attention_in_vit}

The use of patches as tokens in Butterfly ViT induces local effect on the division of blocks. This depends on the size of block to attend and the stride value to jump between tokens. This is depicted by the Figure~\ref{fig:butterfly_in_patch_image}. This structure might also help the sparse linear and non-linear MLP-Mixer architectures as shown in Table~\ref{tab:butterfly_in_mixer}.

We randomize input patches to remove the inductive bias of locality in Butterfly ViT and only test for structured sparse mixing. We find that random tokens still perform well as shown in Table~\ref{tab:sparse_attention_random}. These experiments do not use positional encoding, hence the butterfly provides even better inductive bias of the locality for processing the images or sequences.

To our surprise - using butterfly sparse attention with random token shuffling is still better than dense attention in some cases. We guess that sparse mixing itself can act as inductive bias.

\section{Permutation of Dimension}

Previous works on Butterfly Structure of Linear Transform approximate matrices with power of 2 (\textit{PoT})~\cite{dao2019learning} using $2\times2$ blocks or using resized \textit{PoT} blocks~\cite{chen2021pixelated}. Deformable Butterfly~\cite{lin2021deformable} generalizes this structure to more flexible sized matrices by using non-\textit{PoT} butterfly factors while Monarch Butterfly~\cite{dao2022monarch} generalizes to two block-diagonal matrices. Our work generalizes butterfly structure to work on square matrices of \textit{multiple of M} rather than \textit{power of M} using only $M\times M$ blocks. Modern architectures use $N \to N$ transforms widely and on multiple devices. Hence it is significant to generalize butterfly structure that consider the size of the blocks. 

Below, we show how our method of butterfly structure is different from Deformable Butterfly when fixing the block size. We use formulation from their work to show the difference.

\textit{Recap on Deformable Butterfly (DeBut):}
The authors define the notion of a real-valued DeBut factor as $R^{(p,q)}_{(r,s,t)} \in \mathbb{R}^{p \times q}$ that contains block matrices along its main diagonal, wherein each block matrix is further partitioned into $r \times s$ blocks of $t \times t$ diagonal matrices.

In essence, such densification flow can be generalized to deformable blocks arising
from the product of two contiguous DeBut factors, one with diagonal sub-blocks $(t > 1)$ and another with dense sub-blocks $(t = 1)$, say $\mathbf{R}^{(p_2,q_2)}_{(r_2,s_2,t_2)} \mathbf{R}^{(p_1,q_1)}_{(r_1,s_1,1)}$
 where $t2 > 1$. It can be further shown that
such densifying product mandates $q_2 = p_1$ and $t_2 = r_1$, leading to:
$$R^{(p_2,q_1)}_{(r_2r_1,s_2s_1,1)} \leftarrow R^{(p_2,p_1)}_{(r_2,s_2,r_1)} R^{(p_1,q_1)}_{(r_1,s_1,1)}$$
\textit{Edge case:}
Let's try to decompose $8\times8$ matrix into two butterfly matrices. 

We may take $q_1=p_1=q_2=p_2=8$, $r_1 = s_1= 4$, $r_2= s_2 = 2$ and $t_2 = 4 (= r_1)$ as per the definition. Here, the block-diagonal matrix have size of 4x4 and 2x2 respectively.

However, if we want to use both block-diagonal matrix of size $4\times4$, then we are limited by their statement that densifying mandates $t_2 = r_1$. 

\textit{With our method:}
Our method does not follow the strict requirement stated previously that $t_2 = r_1$, but follows requirements that $q_2 = p_1$, $t_2 > 1$ and $t_2 \leq r_1 $ leading to:
$$R^{(p_2,q_1)}_{(r_2t_2,s_2t_2,1)} \leftarrow R^{(p_2,p_1)}_{(r_2,s_2,t_2)} R^{(p_1,q_1)}_{(r_1,s_1,1)}$$ 
Take $q_1=p_1=q_2=p_2=8$, $r_1 = s_1 = r_2 = s_2 = 4$ and $t_2 = 2 (\neq r_1)$. 
Here, we want to process $8 \times 8$ dimensions with block-diagonal mixers of size $4 \times 4$. 
This can approximate $8\times8$ matrix with two layers of $4\times4$ block diagonal matrices. We achieve this by using stride($t_2$) $=q_2 / r_1$ in the last layer as shown in Algorithm~\ref{algo:butterfly_mlp}~\ref{algo:butterfly_attention}.

\section{More usage of Dimension Group Mixing}
We show further use of partial or group mixing to create more parallelizable and efficient architecture.

\subsection{Token Parallel Attention}

\begin{figure*}[h]
    \centering
    \includegraphics[width=1.0\linewidth]{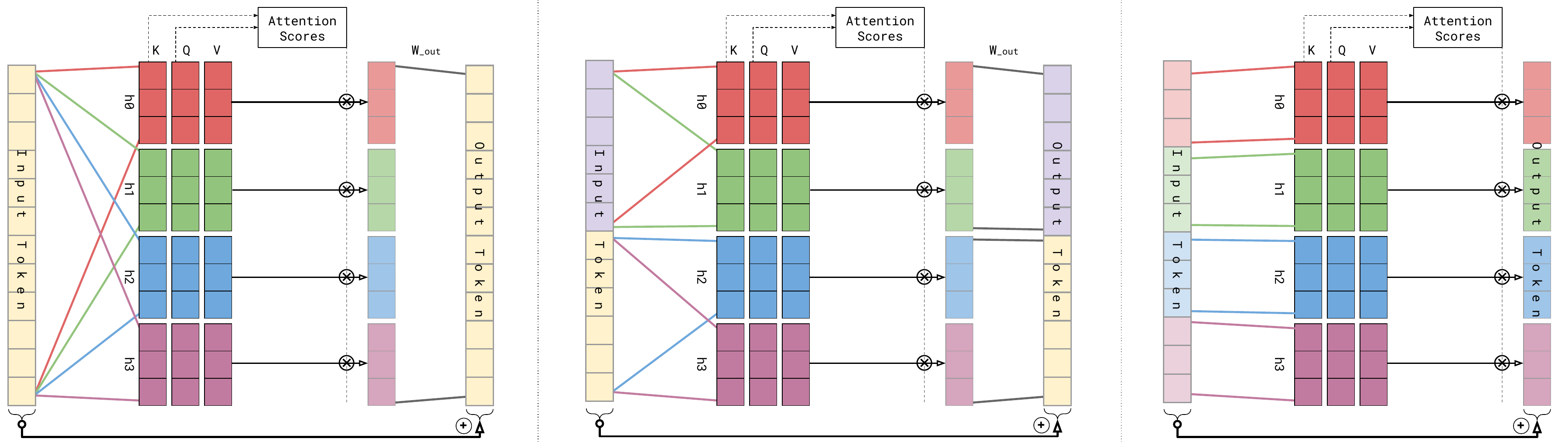} 
    
    \caption{
    ($left$) The standard multi-headed self attention (MHSA). ($mid$) Token group parallel attention with token split into multiple parallel but smaller MHSA ($right$) Most reduced form that uses 1 head per token-group and token dimension is mixed by the next MLP layer. The attention heads in all figures are labed as $h_i$.}
    \label{fig:token_mixer}
\end{figure*}

\begin{table}[h]
    \centering
    \caption{
    The number of heads is 8 for all experiments. Token-Parallel-2 has 4 parallel MultiHeaded Self-Attention  with 2 heads per group. Token-Parallel 1 has 1 Self-Attention per group without using the $W_{out}$ matrix. The baseline is reported from Table~\ref{tab:sparse_attention}. The reported accuracy is best among 3 runs.}
    \resizebox{\linewidth}{!}{
        \begin{tabular}{@{}rccccccc@{}}

\toprule 

Layers, Patch & \multicolumn{2}{c}{8, 4} & \multicolumn{2}{c}{4, 4} & \multicolumn{2}{c}{4, 2}\\
Token-Dims  & \multicolumn{2}{c}{128} & \multicolumn{2}{c}{128} & \multicolumn{2}{c}{64}\\
\cmidrule(lr){2-3} \cmidrule(lr){4-5} \cmidrule(lr){6-7}

& Acc & Params & Acc & Params & Acc & Params \\

\midrule

Baseline & 81.90 & 1.14 M & 81.04 & 0.618 M & 78.35 & 0.299 M \\
without $W_{out}$ & 81.02 & 1.06 M & 80.74 & 0.552 M & 77.89 & 0.282 M \\
Token-Parallel 2 & 81.57  & 0.721 M & 81.42 & 0.405 M & 76.95 & 0.245 M\\
Token-Parallel 1 & 82.18  & 0.689 M & 81.59 & 0.389 M& 76.83 & 0.241 M \\

\bottomrule

\end{tabular}

    }
    \label{tab:sparse_token_attention}
\end{table}

We focus on $W_{out}$ term of attention. First, according to our theory, its purpose is to mix multiple heads.
It is clear that MLP layer just after attention can mix those signals. Can we simply remove that ? The KQV matrices also take into consideration the whole of tokens. So, can we have block-sparse KQV matrices rather than low-rank for each heads in parallel? There has been some success with using split channel for reversible neural architectures~\cite{sukthanker2022generative, mangalam2022reversible}.

We experiment on ViT with 2 heads per parallel-attention-block as shown in Figure~\ref{fig:token_mixer} and find reduced parameter and compute with not much degradation in performance.

On top of that, if we have 1 head per parallel block, we can simply remove the $W_{out}$ matrix algebraically, as it can be combined with $W_v$ matrix. Experiments show that, it works fine and additionally helps fully parallelize heads in multi-headed self attention.

\subsection{Convolutional Channel Mixer}

\begin{figure*}[h]
    \centering
    \includegraphics[angle=90, width=1.0\linewidth]{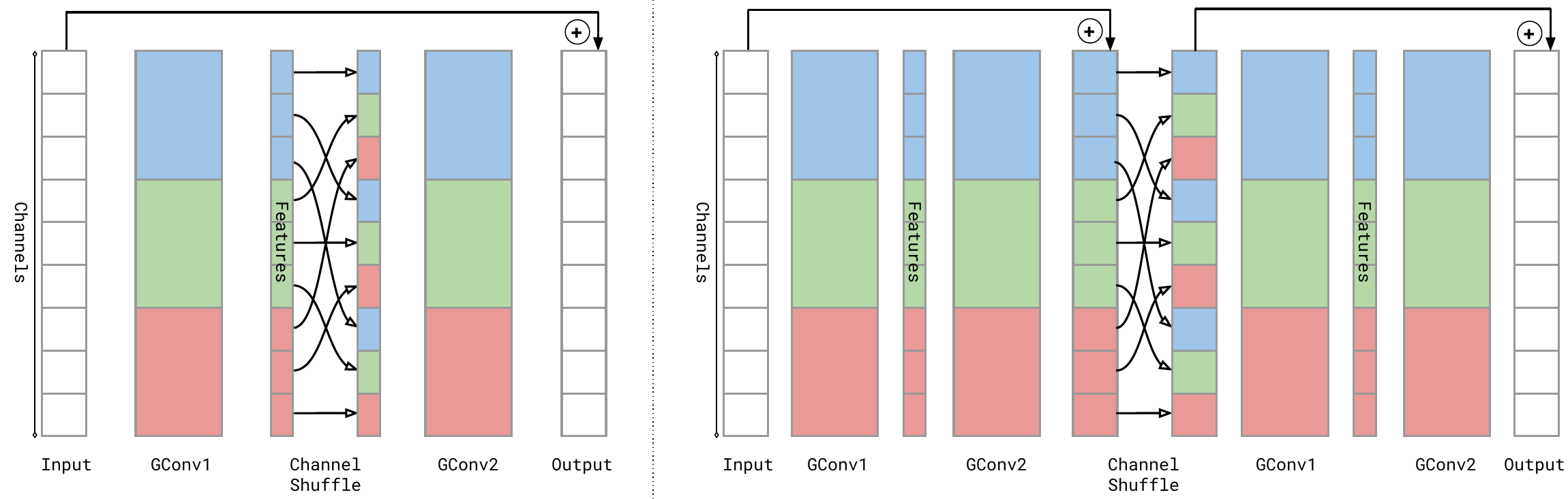} 
    
    \caption{(LEFT) Residual block with Group Convolution and Channel Shuffle for mixing all signals in 1 Res-block. (RIGHT) Multiple grouped residual block with channel shuffle after residual block to mix channels using multiple residual-blocks. GConv1 layer consists of convolution, batchnorm and activation function, whereas Gconv2 consists of convolution and batchnorm.}
    \label{fig:convolution_mixer_archi}
\end{figure*}

\begin{figure}[h]
    \centering
    \includegraphics[width=1.0\linewidth]{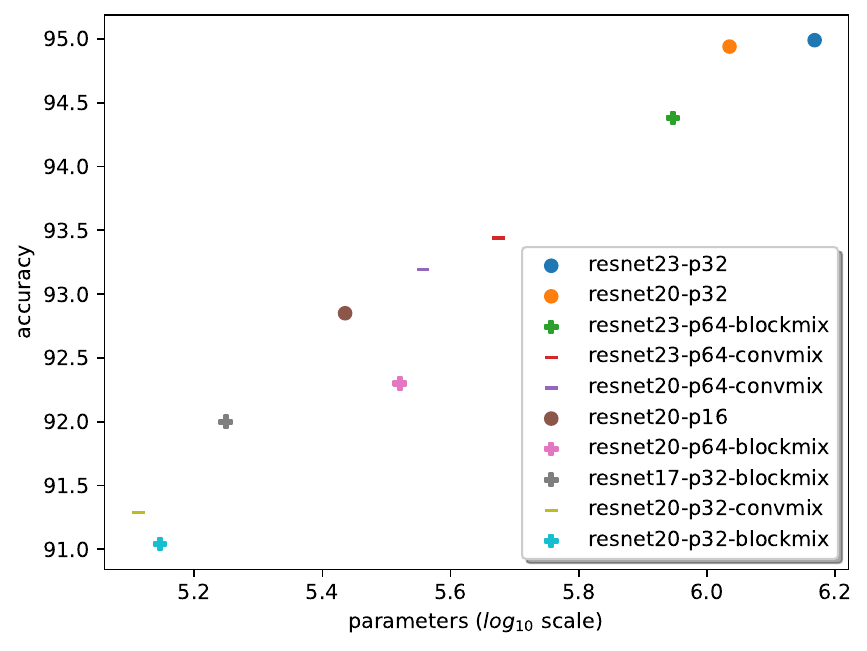} 
    \caption{Result showing test accuracy of various ResNet modifications using different number of input-planes or chanels. The legend shows resnet\{depth\}-p\{input channels\}-\{mixing location\}. conv-mix and block-mix is shown in Fig~\ref{fig:convolution_mixer_archi} left and right respectively.The reported accuracy is best over 3 runs.}
    \label{fig:convolution_mixer_result}
\end{figure}

ShuffelNet~\cite{zhang2018shufflenet} was one of the initial motivations which used grouped convolution, channel shuffeling and grouped convolution in sequence to mix all channels. It focuses on sparse convnet using DepthWise~\cite{chollet2017xception} and PointWise convolution. For experimental comparision on CIFAR-10 dataset, we sparsify the convolution of ResNet~\cite{he2016deep} architecture to mix within a single block as shown in Figure~\ref{fig:convolution_mixer_archi} (left).

One extension we make is to create parallel residual blocks grouped in channels, which do not mix all channels. We shuffle the groups and mix using next layer of parallel res-blocks. Extending this way helps us mix all signals with fewer shuffels/permutation for a constant number of layers as shown in Figure~\ref{fig:convolution_mixer_archi} (right). 

With experiments as shown in Figure~\ref{fig:convolution_mixer_result}, we find that splitting the channels does not help much with sparsity-accuracy tradeoff. The results points to having accuracy dependent on parameters, and not improving due to sparsity. This could be because there is no structure (or inductive bias) to benefit the performance. This approach might be helpful in creating parallel architectures, which can leverage fewer communication between blocks instead of communicating after each convolution operation.

}

{
\bibliography{biblography}

\begin{thebibliography}{10}
\providecommand{\url}[1]{#1}
\csname url@samestyle\endcsname
\providecommand{\newblock}{\relax}
\providecommand{\bibinfo}[2]{#2}
\providecommand{\BIBentrySTDinterwordspacing}{\spaceskip=0pt\relax}
\providecommand{\BIBentryALTinterwordstretchfactor}{4}
\providecommand{\BIBentryALTinterwordspacing}{\spaceskip=\fontdimen2\font plus
\BIBentryALTinterwordstretchfactor\fontdimen3\font minus \fontdimen4\font\relax}
\providecommand{\BIBforeignlanguage}[2]{{%
\expandafter\ifx\csname l@#1\endcsname\relax
\typeout{** WARNING: IEEEtran.bst: No hyphenation pattern has been}%
\typeout{** loaded for the language `#1'. Using the pattern for}%
\typeout{** the default language instead.}%
\else
\language=\csname l@#1\endcsname
\fi
#2}}
\providecommand{\BIBdecl}{\relax}
\BIBdecl

\bibitem{lecun2015lenet}
Y.~LeCun \emph{et~al.}, ``Lenet-5, convolutional neural networks,'' \emph{URL: http://yann. lecun. com/exdb/lenet}, vol.~20, no.~5, p.~14, 2015.

\bibitem{krizhevsky2012imagenet}
A.~Krizhevsky, I.~Sutskever, and G.~E. Hinton, ``Imagenet classification with deep convolutional neural networks,'' \emph{Advances in neural information processing systems}, vol.~25, pp. 1097--1105, 2012.

\bibitem{simonyan2014very}
K.~Simonyan and A.~Zisserman, ``Very deep convolutional networks for large-scale image recognition,'' \emph{arXiv preprint arXiv:1409.1556}, 2014.

\bibitem{he2016deep}
K.~He, X.~Zhang, S.~Ren, and J.~Sun, ``Deep residual learning for image recognition,'' in \emph{Proceedings of the IEEE conference on computer vision and pattern recognition}, 2016, pp. 770--778.

\bibitem{vaswani2017attention}
A.~Vaswani, N.~Shazeer, N.~Parmar, J.~Uszkoreit, L.~Jones, A.~N. Gomez, {\L}.~Kaiser, and I.~Polosukhin, ``Attention is all you need,'' \emph{Advances in neural information processing systems}, vol.~30, 2017.

\bibitem{dosovitskiy2020image}
A.~Dosovitskiy, L.~Beyer, A.~Kolesnikov, D.~Weissenborn, X.~Zhai, T.~Unterthiner, M.~Dehghani, M.~Minderer, G.~Heigold, S.~Gelly \emph{et~al.}, ``An image is worth 16x16 words: Transformers for image recognition at scale,'' \emph{arXiv preprint arXiv:2010.11929}, 2020.

\bibitem{tolstikhin2021mlp}
I.~O. Tolstikhin, N.~Houlsby, A.~Kolesnikov, L.~Beyer, X.~Zhai, T.~Unterthiner, J.~Yung, A.~Steiner, D.~Keysers, J.~Uszkoreit \emph{et~al.}, ``Mlp-mixer: An all-mlp architecture for vision,'' \emph{Advances in Neural Information Processing Systems}, vol.~34, pp. 24\,261--24\,272, 2021.

\bibitem{dinh2014nice}
L.~Dinh, D.~Krueger, and Y.~Bengio, ``Nice: Non-linear independent components estimation,'' \emph{arXiv preprint arXiv:1410.8516}, 2014.

\bibitem{gomez2017reversible}
A.~N. Gomez, M.~Ren, R.~Urtasun, and R.~B. Grosse, ``The reversible residual network: Backpropagation without storing activations,'' \emph{Advances in neural information processing systems}, vol.~30, 2017.

\bibitem{zhang2018shufflenet}
X.~Zhang, X.~Zhou, M.~Lin, and J.~Sun, ``Shufflenet: An extremely efficient convolutional neural network for mobile devices,'' in \emph{Proceedings of the IEEE conference on computer vision and pattern recognition}, 2018, pp. 6848--6856.

\bibitem{ma2018shufflenet}
N.~Ma, X.~Zhang, H.-T. Zheng, and J.~Sun, ``Shufflenet v2: Practical guidelines for efficient cnn architecture design,'' in \emph{Proceedings of the European conference on computer vision (ECCV)}, 2018, pp. 116--131.

\bibitem{idelbayev2020low}
Y.~Idelbayev and M.~A. Carreira-Perpin{\'a}n, ``Low-rank compression of neural nets: Learning the rank of each layer,'' in \emph{Proceedings of the IEEE/CVF Conference on Computer Vision and Pattern Recognition}, 2020, pp. 8049--8059.

\bibitem{cooley1965algorithm}
J.~W. Cooley and J.~W. Tukey, ``An algorithm for the machine calculation of complex fourier series,'' \emph{Mathematics of computation}, vol.~19, no.~90, pp. 297--301, 1965.

\bibitem{prabhu2020butterfly}
A.~Prabhu, A.~Farhadi, M.~Rastegari \emph{et~al.}, ``Butterfly transform: An efficient fft based neural architecture design,'' in \emph{Proceedings of the IEEE/CVF Conference on Computer Vision and Pattern Recognition}, 2020, pp. 12\,024--12\,033.

\bibitem{chen2021pixelated}
B.~Chen, T.~Dao, K.~Liang, J.~Yang, Z.~Song, A.~Rudra, and C.~Re, ``Pixelated butterfly: Simple and efficient sparse training for neural network models,'' \emph{arXiv preprint arXiv:2112.00029}, 2021.

\bibitem{dao2022monarch}
T.~Dao, B.~Chen, N.~S. Sohoni, A.~Desai, M.~Poli, J.~Grogan, A.~Liu, A.~Rao, A.~Rudra, and C.~R{\'e}, ``Monarch: Expressive structured matrices for efficient and accurate training,'' in \emph{International Conference on Machine Learning}.\hskip 1em plus 0.5em minus 0.4em\relax PMLR, 2022, pp. 4690--4721.

\bibitem{liu2021swin}
Z.~Liu, Y.~Lin, Y.~Cao, H.~Hu, Y.~Wei, Z.~Zhang, S.~Lin, and B.~Guo, ``Swin transformer: Hierarchical vision transformer using shifted windows,'' in \emph{Proceedings of the IEEE/CVF International Conference on Computer Vision}, 2021, pp. 10\,012--10\,022.

\bibitem{chen2021regionvit}
C.-F. Chen, R.~Panda, and Q.~Fan, ``Regionvit: Regional-to-local attention for vision transformers,'' \emph{arXiv preprint arXiv:2106.02689}, 2021.

\bibitem{tay2020efficient}
Y.~Tay, M.~Dehghani, D.~Bahri, and D.~Metzler, ``Efficient transformers: A survey,'' \emph{ACM Computing Surveys (CSUR)}, 2020.

\bibitem{tu2022maxvit}
Z.~Tu, H.~Talebi, H.~Zhang, F.~Yang, P.~Milanfar, A.~Bovik, and Y.~Li, ``Maxvit: Multi-axis vision transformer,'' \emph{arXiv preprint arXiv:2204.01697}, 2022.

\bibitem{hassani2022neighborhood}
A.~Hassani, S.~Walton, J.~Li, S.~Li, and H.~Shi, ``Neighborhood attention transformer,'' \emph{arXiv preprint arXiv:2204.07143}, 2022.

\bibitem{chen2021scatterbrain}
B.~Chen, T.~Dao, E.~Winsor, Z.~Song, A.~Rudra, and C.~R{\'e}, ``Scatterbrain: Unifying sparse and low-rank attention,'' \emph{Advances in Neural Information Processing Systems}, vol.~34, pp. 17\,413--17\,426, 2021.

\bibitem{chen2022vit}
B.~Chen, R.~Wang, D.~Ming, and X.~Feng, ``Vit-p: Rethinking data-efficient vision transformers from locality,'' \emph{arXiv preprint arXiv:2203.02358}, 2022.

\bibitem{beltagy2020longformer}
I.~Beltagy, M.~E. Peters, and A.~Cohan, ``Longformer: The long-document transformer,'' \emph{arXiv preprint arXiv:2004.05150}, 2020.

\bibitem{child2019generating}
R.~Child, S.~Gray, A.~Radford, and I.~Sutskever, ``Generating long sequences with sparse transformers,'' \emph{arXiv preprint arXiv:1904.10509}, 2019.

\bibitem{correia2019adaptively}
G.~M. Correia, V.~Niculae, and A.~F. Martins, ``Adaptively sparse transformers,'' \emph{arXiv preprint arXiv:1909.00015}, 2019.

\bibitem{xiong2021nystromformer}
Y.~Xiong, Z.~Zeng, R.~Chakraborty, M.~Tan, G.~Fung, Y.~Li, and V.~Singh, ``Nystr{\"o}mformer: A nystr{\"o}m-based algorithm for approximating self-attention,'' in \emph{Proceedings of the AAAI Conference on Artificial Intelligence}, vol.~35, 2021, pp. 14\,138--14\,148.

\bibitem{kitaev2020reformer}
N.~Kitaev, {\L}.~Kaiser, and A.~Levskaya, ``Reformer: The efficient transformer,'' \emph{arXiv preprint arXiv:2001.04451}, 2020.

\bibitem{wang2020linformer}
S.~Wang, B.~Z. Li, M.~Khabsa, H.~Fang, and H.~Ma, ``Linformer: Self-attention with linear complexity,'' \emph{arXiv preprint arXiv:2006.04768}, 2020.

\bibitem{choromanski2020rethinking}
K.~Choromanski, V.~Likhosherstov, D.~Dohan, X.~Song, A.~Gane, T.~Sarlos, P.~Hawkins, J.~Davis, A.~Mohiuddin, L.~Kaiser \emph{et~al.}, ``Rethinking attention with performers,'' \emph{arXiv preprint arXiv:2009.14794}, 2020.

\bibitem{tay2021synthesizer}
Y.~Tay, D.~Bahri, D.~Metzler, D.-C. Juan, Z.~Zhao, and C.~Zheng, ``Synthesizer: Rethinking self-attention for transformer models,'' in \emph{International conference on machine learning}.\hskip 1em plus 0.5em minus 0.4em\relax PMLR, 2021, pp. 10\,183--10\,192.

\bibitem{wang2020cluster}
S.~Wang, L.~Zhou, Z.~Gan, Y.-C. Chen, Y.~Fang, S.~Sun, Y.~Cheng, and J.~Liu, ``Cluster-former: Clustering-based sparse transformer for long-range dependency encoding,'' \emph{arXiv preprint arXiv:2009.06097}, 2020.

\bibitem{fedus2021switch}
W.~Fedus, B.~Zoph, and N.~Shazeer, ``Switch transformers: Scaling to trillion parameter models with simple and efficient sparsity,'' 2021.

\bibitem{liu2022we}
R.~Liu, Y.~Li, L.~Tao, D.~Liang, and H.-T. Zheng, ``Are we ready for a new paradigm shift? a survey on visual deep mlp,'' \emph{Patterns}, vol.~3, no.~7, p. 100520, 2022.

\bibitem{tang2022image}
Y.~Tang, K.~Han, J.~Guo, C.~Xu, Y.~Li, C.~Xu, and Y.~Wang, ``An image patch is a wave: Phase-aware vision mlp,'' in \emph{Proceedings of the IEEE/CVF Conference on Computer Vision and Pattern Recognition}, 2022, pp. 10\,935--10\,944.

\bibitem{chen2021cyclemlp}
S.~Chen, E.~Xie, C.~Ge, D.~Liang, and P.~Luo, ``Cyclemlp: A mlp-like architecture for dense prediction,'' \emph{arXiv preprint arXiv:2107.10224}, 2021.

\bibitem{wang2022dynamixer}
Z.~Wang, W.~Jiang, Y.~M. Zhu, L.~Yuan, Y.~Song, and W.~Liu, ``Dynamixer: a vision mlp architecture with dynamic mixing,'' in \emph{International Conference on Machine Learning}.\hskip 1em plus 0.5em minus 0.4em\relax PMLR, 2022, pp. 22\,691--22\,701.

\bibitem{guo2022hire}
J.~Guo, Y.~Tang, K.~Han, X.~Chen, H.~Wu, C.~Xu, C.~Xu, and Y.~Wang, ``Hire-mlp: Vision mlp via hierarchical rearrangement,'' in \emph{Proceedings of the IEEE/CVF Conference on Computer Vision and Pattern Recognition}, 2022, pp. 826--836.

\bibitem{zheng2022mixing}
H.~Zheng, P.~He, W.~Chen, and M.~Zhou, ``Mixing and shifting: Exploiting global and local dependencies in vision mlps,'' \emph{arXiv preprint arXiv:2202.06510}, 2022.

\bibitem{yu2021s}
T.~Yu, X.~Li, Y.~Cai, M.~Sun, and P.~Li, ``S2-mlpv2: Improved spatial-shift mlp architecture for vision,'' \emph{arXiv preprint arXiv:2108.01072}, 2021.

\bibitem{hou2022vision}
Q.~Hou, Z.~Jiang, L.~Yuan, M.-M. Cheng, S.~Yan, and J.~Feng, ``Vision permutator: A permutable mlp-like architecture for visual recognition,'' \emph{IEEE Transactions on Pattern Analysis and Machine Intelligence}, 2022.

\bibitem{liu2021pay}
H.~Liu, Z.~Dai, D.~So, and Q.~V. Le, ``Pay attention to mlps,'' \emph{Advances in Neural Information Processing Systems}, vol.~34, pp. 9204--9215, 2021.

\bibitem{lian2021mlp}
D.~Lian, Z.~Yu, X.~Sun, and S.~Gao, ``As-mlp: An axial shifted mlp architecture for vision,'' \emph{arXiv preprint arXiv:2107.08391}, 2021.

\bibitem{shoeybi2019megatron}
M.~Shoeybi, M.~Patwary, R.~Puri, P.~LeGresley, J.~Casper, and B.~Catanzaro, ``Megatron-lm: Training multi-billion parameter language models using model parallelism,'' \emph{arXiv preprint arXiv:1909.08053}, 2019.

\bibitem{kingma2018glow}
D.~P. Kingma and P.~Dhariwal, ``Glow: Generative flow with invertible 1x1 convolutions,'' \emph{Advances in neural information processing systems}, vol.~31, 2018.

\bibitem{jacobsen2018revnet}
J.-H. Jacobsen, A.~Smeulders, and E.~Oyallon, ``i-revnet: Deep invertible networks,'' \emph{arXiv preprint arXiv:1802.07088}, 2018.

\bibitem{ho2019flow++}
J.~Ho, X.~Chen, A.~Srinivas, Y.~Duan, and P.~Abbeel, ``Flow++: Improving flow-based generative models with variational dequantization and architecture design,'' in \emph{International Conference on Machine Learning}.\hskip 1em plus 0.5em minus 0.4em\relax PMLR, 2019, pp. 2722--2730.

\bibitem{teshima2020coupling}
T.~Teshima, I.~Ishikawa, K.~Tojo, K.~Oono, M.~Ikeda, and M.~Sugiyama, ``Coupling-based invertible neural networks are universal diffeomorphism approximators,'' \emph{Advances in Neural Information Processing Systems}, vol.~33, pp. 3362--3373, 2020.

\bibitem{fawzi2022discovering}
A.~Fawzi, M.~Balog, A.~Huang, T.~Hubert, B.~Romera-Paredes, M.~Barekatain, A.~Novikov, F.~J. R~Ruiz, J.~Schrittwieser, G.~Swirszcz \emph{et~al.}, ``Discovering faster matrix multiplication algorithms with reinforcement learning,'' \emph{Nature}, vol. 610, no. 7930, pp. 47--53, 2022.

\bibitem{strassen1969gaussian}
V.~Strassen \emph{et~al.}, ``Gaussian elimination is not optimal,'' \emph{Numerische mathematik}, vol.~13, no.~4, pp. 354--356, 1969.

\bibitem{peng2017large}
C.~Peng, X.~Zhang, G.~Yu, G.~Luo, and J.~Sun, ``Large kernel matters--improve semantic segmentation by global convolutional network,'' in \emph{Proceedings of the IEEE conference on computer vision and pattern recognition}, 2017, pp. 4353--4361.

\bibitem{liu2022more}
S.~Liu, T.~Chen, X.~Chen, X.~Chen, Q.~Xiao, B.~Wu, M.~Pechenizkiy, D.~Mocanu, and Z.~Wang, ``More convnets in the 2020s: Scaling up kernels beyond 51x51 using sparsity,'' \emph{arXiv preprint arXiv:2207.03620}, 2022.

\bibitem{tan2021efficientnetv2}
M.~Tan and Q.~Le, ``Efficientnetv2: Smaller models and faster training,'' in \emph{International Conference on Machine Learning}.\hskip 1em plus 0.5em minus 0.4em\relax PMLR, 2021, pp. 10\,096--10\,106.

\bibitem{howard2017mobilenets}
A.~G. Howard, M.~Zhu, B.~Chen, D.~Kalenichenko, W.~Wang, T.~Weyand, M.~Andreetto, and H.~Adam, ``Mobilenets: Efficient convolutional neural networks for mobile vision applications,'' \emph{arXiv preprint arXiv:1704.04861}, 2017.

\bibitem{han2015learning}
S.~Han, J.~Pool, J.~Tran, and W.~Dally, ``Learning both weights and connections for efficient neural network,'' \emph{Advances in neural information processing systems}, vol.~28, 2015.

\bibitem{gale2020sparse}
T.~Gale, M.~Zaharia, C.~Young, and E.~Elsen, ``Sparse gpu kernels for deep learning,'' in \emph{SC20: International Conference for High Performance Computing, Networking, Storage and Analysis}.\hskip 1em plus 0.5em minus 0.4em\relax IEEE, 2020, pp. 1--14.

\bibitem{dao2019learning}
T.~Dao, A.~Gu, M.~Eichhorn, A.~Rudra, and C.~R{\'e}, ``Learning fast algorithms for linear transforms using butterfly factorizations,'' in \emph{International conference on machine learning}.\hskip 1em plus 0.5em minus 0.4em\relax PMLR, 2019, pp. 1517--1527.

\bibitem{fan2022adaptable}
H.~Fan, T.~Chau, S.~I. Venieris, R.~Lee, A.~Kouris, W.~Luk, N.~D. Lane, and M.~S. Abdelfattah, ``Adaptable butterfly accelerator for attention-based nns via hardware and algorithm co-design,'' in \emph{2022 55th IEEE/ACM International Symposium on Microarchitecture (MICRO)}.\hskip 1em plus 0.5em minus 0.4em\relax IEEE, 2022, pp. 599--615.

\bibitem{tay2020long}
Y.~Tay, M.~Dehghani, S.~Abnar, Y.~Shen, D.~Bahri, P.~Pham, J.~Rao, L.~Yang, S.~Ruder, and D.~Metzler, ``Long range arena: A benchmark for efficient transformers,'' \emph{arXiv preprint arXiv:2011.04006}, 2020.

\bibitem{nangia2018listops}
N.~Nangia and S.~R. Bowman, ``Listops: A diagnostic dataset for latent tree learning,'' \emph{arXiv preprint arXiv:1804.06028}, 2018.

\bibitem{maas2011learning}
A.~Maas, R.~E. Daly, P.~T. Pham, D.~Huang, A.~Y. Ng, and C.~Potts, ``Learning word vectors for sentiment analysis,'' in \emph{Proceedings of the 49th annual meeting of the association for computational linguistics: Human language technologies}, 2011, pp. 142--150.

\bibitem{radev2013acl}
D.~R. Radev, P.~Muthukrishnan, V.~Qazvinian, and A.~Abu-Jbara, ``The acl anthology network corpus,'' \emph{Language Resources and Evaluation}, vol.~47, pp. 919--944, 2013.

\bibitem{krizhevsky2009learning}
A.~Krizhevsky, G.~Hinton \emph{et~al.}, ``Learning multiple layers of features from tiny images,'' 2009.

\bibitem{linsley2018learning}
D.~Linsley, J.~Kim, V.~Veerabadran, C.~Windolf, and T.~Serre, ``Learning long-range spatial dependencies with horizontal gated recurrent units,'' \emph{Advances in neural information processing systems}, vol.~31, 2018.

\bibitem{bronstein2017geometric}
M.~M. Bronstein, J.~Bruna, Y.~LeCun, A.~Szlam, and P.~Vandergheynst, ``Geometric deep learning: going beyond euclidean data,'' \emph{IEEE Signal Processing Magazine}, vol.~34, no.~4, pp. 18--42, 2017.

\bibitem{bronstein2021geometric}
M.~M. Bronstein, J.~Bruna, T.~Cohen, and P.~Veli{\v{c}}kovi{\'c}, ``Geometric deep learning: Grids, groups, graphs, geodesics, and gauges,'' \emph{arXiv preprint arXiv:2104.13478}, 2021.

\bibitem{dao2022flashattention}
T.~Dao, D.~Fu, S.~Ermon, A.~Rudra, and C.~R{\'e}, ``Flashattention: Fast and memory-efficient exact attention with io-awareness,'' \emph{Advances in Neural Information Processing Systems}, vol.~35, pp. 16\,344--16\,359, 2022.

\bibitem{lin2021deformable}
R.~Lin, J.~Ran, K.~H. Chiu, G.~Chesi, and N.~Wong, ``Deformable butterfly: A highly structured and sparse linear transform,'' \emph{Advances in Neural Information Processing Systems}, vol.~34, pp. 16\,145--16\,157, 2021.

\bibitem{hayase2023mlp}
T.~Hayase and R.~Karakida, ``Mlp-mixer as a wide and sparse mlp,'' \emph{arXiv preprint arXiv:2306.01470}, 2023.

\bibitem{deng2009imagenet}
J.~Deng, W.~Dong, R.~Socher, L.-J. Li, K.~Li, and L.~Fei-Fei, ``Imagenet: A large-scale hierarchical image database,'' in \emph{2009 IEEE conference on computer vision and pattern recognition}.\hskip 1em plus 0.5em minus 0.4em\relax Ieee, 2009, pp. 248--255.

\bibitem{sukthanker2022generative}
R.~S. Sukthanker, Z.~Huang, S.~Kumar, R.~Timofte, and L.~Van~Gool, ``Generative flows with invertible attentions,'' in \emph{Proceedings of the IEEE/CVF Conference on Computer Vision and Pattern Recognition}, 2022, pp. 11\,234--11\,243.

\bibitem{mangalam2022reversible}
K.~Mangalam, H.~Fan, Y.~Li, C.-Y. Wu, B.~Xiong, C.~Feichtenhofer, and J.~Malik, ``Reversible vision transformers,'' in \emph{Proceedings of the IEEE/CVF Conference on Computer Vision and Pattern Recognition}, 2022, pp. 10\,830--10\,840.

\bibitem{chollet2017xception}
F.~Chollet, ``Xception: Deep learning with depthwise separable convolutions,'' in \emph{Proceedings of the IEEE conference on computer vision and pattern recognition}, 2017, pp. 1251--1258.

\end{thebibliography}
\bibliographystyle{IEEEtran}
}

\end{document}